\documentclass[12pt]{article}
\usepackage{amsmath,amsfonts}
\usepackage{geometry}
\usepackage{times}
\usepackage{graphicx}
\usepackage{color}
\usepackage{bm}
\usepackage{multirow}
\usepackage[authoryear]{natbib}
\usepackage{rotating}
\usepackage{bbm}
\usepackage{latexsym}
\usepackage{subcaption}
\usepackage{url}

\usepackage{algorithm}
\usepackage[noend]{algpseudocode}

\newcommand{\s}{\bm{s}}
\renewcommand{\r}{\bm{r}}
\newcommand{\vtheta}{\bm{\theta}}
\newcommand{\data}{\mathcal{D}}
\newcommand{\I}{\mathcal{I}}
\renewcommand{\L}{\mathcal{L}}
\newcommand{\x}{\bm{x}}
\newcommand{\Libs}{\hat{\L}_{\text{IBS}}}
\newcommand{\Libsrep}[1]{\hat{\L}_{\text{IBS-}{#1}}}
\newcommand{\lapse}{\gamma}
\newcommand{\thresh}{\xi}

\newcommand{\Reps}{R}
\newcommand{\reps}{r}
\newcommand{\R}{\bm{R}}

\newcommand{\eq}{Equation}
\newcommand{\eqs}{Equations}

\newcommand{\Pchange}{P_{\text{correct}}\left(\Delta_s^{(c)}; \vtheta \right)}
\newcommand{\budget}{S}

\algnewcommand\algorithmicinput{\textbf{Input:}}
\algnewcommand\INPUT{\item[\algorithmicinput]}

\algloop{Do}

\newcommand{\captionfonts}{\normalsize}

\makeatletter  
\long\def\@makecaption#1#2{%
  \vskip\abovecaptionskip
  \sbox\@tempboxa{{\captionfonts #1: #2}}%
  \ifdim \wd\@tempboxa >\hsize
    {\captionfonts #1: #2\par}
  \else
    \hbox to\hsize{\hfil\box\@tempboxa\hfil}%
  \fi
  \vskip\belowcaptionskip}
\makeatother   

\renewcommand{\thefootnote}{\normalsize \arabic{footnote}}

\begin{document}
\hspace{13.9cm}1

\ \vspace{20mm}\\

{\LARGE Unbiased and Efficient Log-Likelihood Estimation with Inverse Binomial Sampling}

\renewcommand*{\thefootnote}{\fnsymbol{footnote}}

\ \\
{\bf \large Bas van Opheusden$^{\displaystyle *, \displaystyle 1, \displaystyle 2}$, Luigi Acerbi$^{\displaystyle *, \displaystyle 1, \displaystyle 3, \displaystyle 4}$, Wei Ji Ma$^{\displaystyle 1, \displaystyle 5}$}\\
{$^{\displaystyle 1}$Center for Neural Science, New York University, New York, NY 10003, U.S.A.}\\
{$^{\displaystyle 2}$Department of Psychology, Princeton University, Princeton, NJ 08540, U.S.A.}\\
{$^{\displaystyle 3}$Department of Computer Science, University of Helsinki, 00560 Helsinki, Finland.}\\
{$^{\displaystyle 4}$Department of Basic Neuroscience, University of Geneva, 1206 Geneva, Switzerland.}\\
{$^{\displaystyle 5}$Department of Psychology, New York University, New York, NY 10003, U.S.A.}\\
{$^{\displaystyle *}$ These authors contributed equally to this work.}\\

\renewcommand*{\thefootnote}{\arabic{footnote}}
\setcounter{footnote}{0}

%


{\bf Corresponding authors:} svo@princeton.edu; luigi.acerbi@helsinki.fi.

\thispagestyle{empty}
\markboth{}{NC instructions}
\ \vspace{-0mm}\\

\clearpage

\begin{center} {\bf Abstract} \end{center}
The fate of scientific hypotheses often relies on the ability of a computational model to explain the data, quantified in modern statistical approaches by the likelihood function. The log-likelihood is the key element for parameter estimation and model evaluation. However, the log-likelihood of complex models in fields such as computational biology and neuroscience is often intractable to compute analytically or numerically. In those cases, researchers can often only estimate the log-likelihood by comparing observed data with synthetic observations generated by model simulations. Standard techniques to approximate the likelihood via simulation either use summary statistics of the data or are at risk of producing substantial biases in the estimate. 
Here, we explore another method, inverse binomial sampling (IBS), which can estimate the log-likelihood of an entire data set efficiently and without bias. For each observation, IBS draws samples from the simulator model until one matches the observation. The log-likelihood estimate is then a function of the number of samples drawn. The variance of this estimator is uniformly bounded, achieves the minimum variance for an unbiased estimator, and we can compute calibrated estimates of the variance.
We provide theoretical arguments in favor of IBS and an empirical assessment of the method for maximum-likelihood estimation with simulation-based models. As case studies, we take three model-fitting problems of increasing complexity from computational and cognitive neuroscience. In all problems, IBS generally produces lower error in the estimated parameters and maximum log-likelihood values than alternative sampling methods with the same average number of samples. Our results demonstrate the potential of IBS as a practical, robust, and easy to implement method for log-likelihood evaluation when exact techniques are not available.

\section{Introduction}
\label{sec:intro}

The \emph{likelihood function} is one of the most important mathematical objects for modern statistical inference. Briefly, the likelihood function measures how well a model with a given set of parameters can explain an observed data set. For a data set of discrete observations, the likelihood has the intuitive interpretation of the probability that a random sample generated from the model matches the data, for a given setting of the model parameters.

In many scientific disciplines, such as computational neuroscience and cognitive science, computational models are used to give a precise quantitative form to scientific hypotheses and theories. Statistical inference then plays at least two fundamental roles for scientific discovery. First, our goal may be \emph{parameter estimation} for a model of interest. Parameter values may have a significance in themselves, for example we may be looking for differences in parameters between distinct experimental conditions in a clinical or behavioral study. Second, we may be considering a number of competing scientific hypotheses, instantiated by different models, and we want to evaluate which model `best' captures the data according to some criteria, such as \emph{explanation} (what evidence the data provide in favor of each model?) and \emph{prediction} (which model best predicts new observations?).

Crucially, the likelihood function is a key element for both parameter estimation and model evaluation.
A principled method to find best-fitting model parameters for a given data set is maximum-likelihood estimation (MLE), which entails optimizing the likelihood function over the parameter space \citep{myung2003tutorial}.  
Other common parameter estimation methods, such as maximum-a-posteriori (MAP) estimation or full or approximate Bayesian inference of posterior distributions, still involve the likelihood function \citep{gelman2013bayesian}.
Moreover, almost all model comparison metrics commonly used for scientific model evaluation are based on likelihood computations, from predictive metrics such as Akaike's information criterion (AIC; \citealp{akaike1974new}), the deviance information criterion (DIC; \citealp{spiegelhalter2002bayesian}), the widely applicable information criterion (WAIC; \citealp{watanabe2010asymptotic}), leave-one-out cross-validation \citep{vehtari2017practical}; to evidence-based metrics such at the marginal likelihood \citep{mackay2003information} and (loose) approximations thereof, such as the Bayesian information criterion (BIC; \citealp{schwarz1978estimating}) or the Laplace approximation \citep{mackay2003information}. 

However, many complex computational models, such as those developed in computational biology \citep{pritchard1999population,ratmann2007using,wilkinson2011stochastic}, neuroscience \citep{pospischil2008minimal,sterratt2011principles} and cognitive science \citep{van2016people}, take the form of a generative model or \emph{simulator}, that is an algorithm which given some context information and parameter settings returns one or more simulated observations (a synthetic data set). 
In those cases, the likelihood is often impossible to calculate analytically, and even when the likelihood might be available in theory, the numerical calculations needed to obtain it might be overwhelmingly expensive and intractable in practice. In such situations, the only thing one can do is to run the model to simulate observations (`samples'). 
In the absence of a likelihood function, common approaches to `likelihood-free inference' generally try and match summary statistics of the data with summary statistics of simulated observations \citep{beaumont2002approximate,wood2010statistical}.


In this paper, we ask instead the question of whether we can use samples from a simulator model to \emph{directly} estimate the likelihood of the full data set, without recurring to summary statistics, in a `correct' and `efficient' manner, for some specific definition of these terms. 
The answer is \emph{yes}, as long as we use the right \emph{sampling method}. 

In brief, a sampling method consists of a `sampling policy' (a rule that determines how long to keep drawing samples for) and an `estimator' which converts the samples to a real-valued number. To estimate the likelihood of a single observation (e.g., the response of a participant on a single trial of a behavioral experiment), the most obvious sampling policy is to draw a fixed amount of samples from the simulator model, and the simplest estimator is the fraction of samples that match the observation (or is `close enough' to it, for continuous observations).
However, most basic applications, such as computing the likelihood of multiple observations, require one to estimate the logarithm of the likelihood, or log-likelihood (see Section~\ref{sec:reduction} for the underlying technical reasons). The `fixed sampling' method described above cannot provide unbiased estimates for the log-likelihood (see Section~\ref{sec:fixed_fail}). Such bias vanishes in the asymptotic limit of infinite samples, but drawing samples from the model can be computationally expensive, especially if the simulator model is complex. In practice, the bias introduced by any fixed sampling method can translate to considerable biases in estimates of model parameters, or even reverse the outcome of model comparison analyses. In other words, using poor sampling methods can cause researchers to draw conclusions about scientific hypotheses which are not supported by their data. 

In this work, we introduce \emph{inverse binomial sampling} (IBS) as a powerful and simple technique for correctly and efficiently estimating log-likelihoods of simulator-based models. Crucially, IBS is a sampling method that provides uniformly unbiased estimates of the log-likelihood~\citep{haldane1945method,degroot1959unbiased} and calibrated estimates of their variance, which is also uniformly bounded. 

We note that the problem of estimating functions $f(p)$ from observations of a Bernoulli distribution with parameter $p$ has been studied for mostly theoretical reasons in the mid-$20$th century, with major contributions from \citet{haldane1945labour,haldane1945method}, \citet{girshick1946unbiased}, \citet{dawson1953unbiased} and \citet{degroot1959unbiased}. These works have largely focused on deriving the set of functions $f(p)$ for which an unbiased estimate exists, and demonstrating that for those functions, the inverse sampling policy (see Section~\ref{sec:policies}) is in a precise sense `efficient'. 
Our main contribution here is to demonstrate that inverse binomial sampling provides a practically and theoretically efficient solution for a common problem in computational modeling; namely likelihood-free inference of complex models. To back up our claims, we provide theoretical arguments for the efficiency of IBS and a practical demonstration of its value for log-likelihood estimation and fitting of simulation-based models, in particular those used in computational cognitive science. We note that \citet{duncan2004unbiased} had previously proposed inverse binomial sampling as a method for likelihood-free inference for certain econometric models, but did not present an empirical assessment of the quality of the estimation and to our knowledge has not led to further adoption of IBS.

The paper is structured as follows. After setting the stage with useful definitions and notation (Section \ref{sec:defs}), we describe more in detail the issues with the fixed sampling method and why they cannot be fixed (Section \ref{sec:fixed_fail}). We then present a series of arguments for why IBS solves these issues, and in particular why being unbiased here is of particular relevance (Section \ref{sec:ibs_better}). 
Then, we present an empirical comparison of IBS and fixed sampling in the setting of maximum-likelihood estimation (Section \ref{sec:experiments}). As case studies, we take three model-fitting problems of increasing complexity from computational cognitive science: an `orientation discrimination' task, a `change localization' task, and a complex sequential decision making task. 
In all problems, IBS generally produces lower error in the estimated parameters than fixed sampling with the same average number of samples. IBS also returns solutions that are very close in value to the true maximum log-likelihood. We conclude by discussing further applications and extensions of IBS (Section \ref{sec:discussion}).
Our theoretical analyses and empirical results demonstrate the potential of IBS as a practical, robust, and easy-to-implement method for log-likelihood evaluation when exact or numerical solutions are unavailable.

Implementations of IBS with tutorials and examples are available at the following link: \url{https://github.com/lacerbi/ibs}.

\section{Definitions and notation}
\label{sec:defs}

The two fundamental ingredients to run IBS are:
\begin{enumerate}
\item A data set $\data=\{(\s_i,\r_i)\}_{i=1}^N$ consisting of $N$ `trials' characterized by `stimuli' $\s_i$ and \emph{discrete} `responses' $\r_i$. 
\item A generative model $g$ for the data (also known as a `simulator'): a stochastic function that takes as input a stimulus $\s$ and a parameter vector $\vtheta$ (and possibly other information) and outputs a response $\r$. 
\end{enumerate}
In this section we expand on and provide motivations for the above assumptions, and introduce related definitions and notation used in the rest of the paper.

Here and in the following, for ease of reference, we use the language of behavioral and cognitive modeling (e.g., `trial' for data points, `stimulus' for independent or contextual variables, `response' for observations or outcomes), but the statistical techniques that we discuss in the paper apply to any model and data set from any domain as long as they satisfy the fundamental assumption of IBS delineated above.

\subsection{The likelihood function}
\label{sec:likelihood}

We assume that we want to model a data set $\data =\{(\s_i,\r_i)\}_{i=1}^N$ consisting of $N$ `trials' (data points), where
\begin{itemize}
\item $\s_i$ is the \emph{stimulus} (i.e., the experimental context, or independent variable) presented on the $i$-th trial; typically, $\s_i$ is a scalar or vector of discrete or continuous variables (more generally, there are no restrictions on what $\s_i$ can be as long as the simulator can accept it as input);
\item $\r_i$ is the \emph{response} (i.e., the experimental observations, outcomes, or dependent variables) measured on the $i$-th trial; $\r_i$ can be a scalar or vector, but crucially we assume it takes \emph{discrete} values.
\end{itemize}
The requirement that $\r_i$ be discrete will be discussed below, in Section \ref{sec:model}.


Given a data set $\data$, and a model parametrized by parameter vector $\vtheta$, we can write the \emph{likelihood function} for the responses given the stimuli and model parameters as
\begin{equation} \label{eq:likelihood_zero}
\begin{split}
\Pr\left(\{\r_i\}_{i=1}^N | \{\s_i\}_{i=1}^N, \vtheta\right)
= & \prod_{i=1}^N \Pr \left(\r_i | \r_1, \ldots, \r_{i-1}, \s_1, \ldots, \s_N,\bm{\theta}\right) \\
= & \prod_{i=1}^N \Pr \left(\r_i | \r_1, \ldots, \r_{i-1}, \s_1, \ldots, \s_i,\bm{\theta}\right), \\
\end{split}
\end{equation}
where the first line follows from the chain rule of probability, and holds in general, whereas in the second step we applied the reasonable `causal' (or `no-time-travel') assumption that the response at the $i$-th trial is not influenced by future stimuli.\footnote{We also used the causality assumption that current responses are not influenced by future responses to choose a specific order to apply the chain rule in the first line.} 

Note that Equation \ref{eq:likelihood_zero} assumes that the researcher is not interested in statistically modeling the stimuli, which are taken to be given (i.e., on the right-hand side of the conditional probability). This choice is without loss of generality, as any variable of statistical interest can always be relabeled and become an element of the `response' vector. For compactness, from now on we will denote $\Pr\left(\{\r_i\}_{i=1}^N | \{\s_i\}_{i=1}^N, \vtheta\right) \equiv \Pr\left(\data | \vtheta\right)$, in a slight abuse of notation.

\eq~\ref{eq:likelihood_zero} describes the most general class of models, in which the response in the current trial might be influenced by the history of both previous stimuli and previous responses. Many models commonly make a stronger conditional independence assumption between trials, such that the response on the current trial only depends on the current stimulus. Under this stronger assumption, the likelihood takes a simpler form,
\begin{equation} \label{eq:likelihood_ind}
\begin{split}
\Pr\left(\data | \vtheta\right) = & \prod_{i=1}^N \Pr \left(\r_i | \s_i,\bm{\theta}\right).
\end{split}
\end{equation}
While \eq~\ref{eq:likelihood_ind} is simpler, it still includes a wide variety of models. For example, note that time-dependence can be easily included in the model by incorporating time into the `stimulus' $\s$, and including time-dependent parameters explicitly in the model specification.
In the rest of this work, for simplicity we consider models that make conditional independence assumptions as in \eq~\ref{eq:likelihood_ind}, but our techniques apply in general also for likelihoods as per \eq~\ref{eq:likelihood_zero}.

Given that the likelihood of the $i$-th trial can be directly interpreted as the probability of observing response $\r_i$ in the $i$-th trial (conditioned on everything else), we denote such quantity with $p_i\in [0,1]$. The value $p_i$ is a function of $\vtheta$, depends on the current stimulus and response, and may or may not depend on previous stimuli or responses. 

With this notation, we can simply write the likelihood as
\begin{equation} \label{eq:likelihood_p}
\begin{split}
\Pr\left(\data | \vtheta\right) = & \prod_{i=1}^N p_i.
\end{split}
\end{equation}


Finally, we note that it is common practice to work with the logarithm of the likelihood, or log-likelihood, that is 
\begin{equation} \label{eq:loglikelihood_p}
\begin{split}
\L(\vtheta) \equiv \log \Pr\left(\data | \vtheta\right) = & \log \prod_{i=1}^N p_i = \sum_{i=1}^N \log p_i.
\end{split}
\end{equation}
The typical rationale for switching to the log-likelihood is that for large $N$ the likelihood tends to be a vanishingly small quantity, so the logarithm makes it easier to handle numerically (`numerical convenience'). However, we will see later (see Section \ref{sec:whynotlikelihood}) that there are statistically meaningful reasons to prefer the logarithmic representation (i.e., a sum of independent terms).


Crucially, we assume that the likelihood function is unavailable in a tractable form -- for example, because the model is too complex to derive an analytical expression for the likelihood. Instead, IBS provides a technique for estimating \eq~\ref{eq:loglikelihood_p} via simulation.

\subsection{The generative model or simulator}
\label{sec:model}

While we assume no availability of an \emph{explicit} representation of the likelihood function, we assume that the model of interest is represented \emph{implicitly} by a stochastic generative model (or `simulator'). 
In the most general case, the simulator is a stochastic function $g$ that takes as input the current stimulus $\s_i$, arrays of past stimuli and responses, and a parameter vector $\vtheta$, and outputs a discrete response $\r_i$, \emph{conditional} on all past events,
\begin{equation} \label{eq:simulator}
\r_i \sim g(\s_1, \ldots, \s_i, \r_1, \ldots, \r_{i-1}; \vtheta).
\end{equation}
As mentioned in the previous section, a common assumption for a model is that the response in the current trial only depends on the current stimulus and parameter vector, in which case
\begin{equation} \label{eq:simulator_ind}
\r \sim g(\s; \vtheta).
\end{equation}
For example, the model $g(\cdot)$ could be simulating the responses of a human participant in a complex cognitive task; the (discrete) choices taken by a rodent in a perceptual decision-making experiment; or the spike count of a neuron in sensory cortex for a specific time bin after a stimulus presentation. 

We list now the requirements that the simulator model needs to satisfy to be used in conjuction with IBS.

\subsubsection*{Discrete response space}

Lacking an expression for the likelihood function, the only way to estimate the likelihood or any function thereof is by drawing samples $\r\sim g\left(\s_i, \ldots; \vtheta \right)$ on each trial, and matching them to the response $\r_i$. 
This approach requires that there is a nonzero probability for a random sample $\r$ to match $\r_i$, hence the assumption that the space of responses is discrete. We will discuss in Section \ref{sec:ABC} a possible method to extend IBS to larger or continuous response spaces. 

\subsubsection*{Conditional simulation}

An important requirement of the generative model, stated implicitly by \eqs~\ref{eq:simulator} and \ref{eq:simulator_ind}, is that the simulator should afford \emph{conditional simulation}, in that we can simulate the response $\r_i$ for any trial $i$, given the current stimulus $\s_i$, and possibly previous stimuli and responses. Note that this class of models, while large, does not include \emph{all} possible simulators, in that some simulators might not afford conditional generation of responses. For example, models with latent dynamics might be able to generate a full sequence of responses given the stimuli, but it might not be easy or computationally tractable to generate the response in a given trial, conditional on a specific sequence of previous responses. 

\subsubsection*{Computational cost}

Finally, for the purpose of some of our analyses we assume that drawing a sample from the generative model is at least moderately computationally expensive, which limits the approximate budget of samples one is willing to use for each likelihood evaluation (in our analyses, up to about a hundred, on average, per likelihood evaluation). Number of samples is a reasonable proxy for any realistic resource expenditure since most costs (e.g., time, energy, number of processors) would be approximately proportional to it. Therefore, we also require that every response value in the data has a non-negligible probability of being sampled from the model -- given the available budget of samples one can reasonably draw. In this paper, we will focus on the low-sample regime, since that is where IBS considerably outperforms other approaches. For our analyses of performance of the algorithm, we also assume that the computational cost is independent of the stimulus, response or model parameters, but this is not a requirement of the method.

\subsection{Reduction to Bernoulli sampling}
\label{sec:reduction}

Given the conditional independence structure codified by \eq~\ref{eq:likelihood_p}, to estimate the log-likelihood of the entire data set, we cannot do better than estimating $p_i$ on each trial independently, and combining the results.
However, combining estimates $\hat{p}_i$ into a well-behaved estimate of $\prod_{i=1}^N p_i$ is non-trivial (see Section \ref{sec:whynotlikelihood}). 
Instead, it is easier to estimate $\L_i\equiv\log p_i$ for each trial and calculate the log-likelihood
\begin{equation}
\L\left(\bm{\theta}\right)=\log \Pr\left(\data|\vtheta\right)=\sum_{i=1}^N\log p_i=\sum_{i=1}^N \L_i.
\end{equation}
We can estimate this log-likelihood by simply summing estimates $\hat{\L}_i$ across trials, in which case the central limit theorem guarantees that the distribution of $\hat{\L}\left(\vtheta\right)$ is normally distributed for large values of $N$, which is true for typical values of $N$ of the order of a hundred or more (see also Section \ref{sec:highermoments}).

We can make one additional simplification, without loss of generality. The generative model specifies an implicit probability distribution 
$\r_i \sim g(\s_i, \dots; \vtheta)$
for each trial. However, to estimate the log-likelihood, we do not need to know the full distribution, only the probability for a random sample $\r$ from the model to match the observed response $\r_i$. Therefore, we can convert each sample $\r$ to 
\begin{equation}
x= \left\{ \begin{array}{cll} 1 &\text{if } \r=\r_i & \text{(`hit')} \\ 0 &\text{otherwise} & \text{(`miss')}, \end{array} \right.
\end{equation}
and lose no information relevant for estimating the log-likelihood. By construction, $x$ follows a Bernoulli distribution with probability $p_i$. Note that this holds regardless of the type of data, the structure of the generative model or the model parameters. The only difference between different models and data sets is the distribution of the likelihood $p_i$ across trials. Moreover, since $p_i$ is interpreted as the parameter of a Bernoulli distribution, we can apply standard frequentist or Bayesian statistical reasoning to it.

In conclusion, we can reduce the problem of estimating the log-likelihood of a given model by sampling to a smaller problem: given a method to draw samples $(x_1,x_2,\dots)$ from a Bernoulli distribution with unknown parameter $p$, estimate $\log p$ as precisely and accurately as possible using on average as few samples as possible. 

\subsection{Sampling policies and estimators}
\label{sec:policies}

A \emph{sampling policy} is a function that, given a sequence of samples $\x \equiv \left(x_1,x_2,\dots, x_k\right)$, decides whether to draw an additional sample or not \citep{girshick1946unbiased}. In this work, we compare two sampling policies:
\begin{enumerate}
\item The commonly used \emph{fixed} policy: Draw a fixed number of samples $M$, then stop. 
\item The \emph{inverse binomial sampling} policy: Keep drawing samples until $x_k=1$, then stop. 
\end{enumerate}

In our case, an \emph{estimator} (of $\log p$) is a function $\hat{\L}(\x)$ that takes as input a sequence of samples $\x = \left(x_1,x_2,\dots, x_k\right)$ and returns an estimate of $\log p$.
We recall that the \emph{bias} of an estimator $\hat{\L}$ of $\log p$, for a given true value of the Bernoulli parameter $p$, is defined as
\begin{equation}
\text{Bias}\left[ \hat{\L} | p \right] = \mathbb{E}\left[\hat{\L}\right]-\log p,
\end{equation}
where the expectation is taken over all possible sequences $\x$ generated by the chosen sampling policy under the Bernoulli probability $p$.
Such estimator is (uniformly) \emph{unbiased} if $\text{Bias}\left[ \hat{\L}| p \right] = 0$ for all $0 < p \le 1$ (that is, the estimator is centered around the true value).

\subsubsection*{Fixed sampling}

For the fixed sampling policy, since all samples are independent and identically distributed, a sufficient statistic for estimating $p$ from the samples $\left(x_1,x_2,\dots, x_M\right)$ is the number of `hits', $m(\x) \equiv \displaystyle\sum_{k=1}^Mx_k$. The most obvious estimator for an obtained sequence of samples $\x$ is then 
\begin{equation} \label{eq:Lbad}
\hat{\L}_{\text{naive}}(\x) =\log\left(\frac{m(\x)}{M}\right),
\end{equation}
but this estimator has infinite bias; since as long as $p\neq 1$, there is always a nonzero chance that $m(\x)=0$, in which case $\hat{\L}_{\text{naive}}(\x) =-\infty$ (and thus $\mathbb{E}\left[ \hat{\L}_{\text{naive}} \right] = -\infty$). This divergence can be fixed in multiple ways; in the main text we use 
\begin{equation}\label{eq:Lfixed}
\hat{\L}_{\text{fixed}}(\x)=\log\left(\frac{m(\x)+1}{M+1}\right).
\end{equation}
Note that \emph{any} estimator based on the fixed sampling policy will always produce biased estimates of $\log p$, as guaranteed by the reasoning in Section~\ref{sec:fixed_fail} below. As an empirical validation, we show in Appendix \ref{sec:ori_details} that our results do not depend on the specific choice of estimator for fixed sampling (Equation \ref{eq:Lfixed}).

\subsubsection*{Inverse binomial sampling}

For inverse binomial sampling we note that, since $x$ is a binary variable, the policy will always result in a sequence of samples of the form 
\begin{equation}
\x = (\overbrace{0,0,0,0,0,\dots,0,1}^{K}),
\end{equation}
where the length of the sequence is a stochastic variable, which we label $K$ (a positive integer).
Moreover, since each sample is independent and a `hit' with probability $p$, the length $K$ follows a geometric distribution with parameter $1-p$, 
\begin{equation}
\Pr\left(K=k\right)=p(1-p)^{k-1}.
\end{equation}
We convert a value of $K$ into an estimate for $\log p$ using the IBS estimator,
\begin{equation}\label{eq:Libs}
\Libs(\x) = \begin{cases} 0 & \text{ for } K = 1\\ -\sum_{k=1}^{K-1}\frac{1}{k} & \text{ for } K > 1. \end{cases}
\end{equation}
Crucially, \eq~\ref{eq:Libs} combined with the IBS policy provides a uniformly unbiased estimator of $\log p$ \citep{degroot1959unbiased}. Moreover, we can show that $\Libs$ is the \emph{uniformly minimum-variance unbiased estimator} of $\log p$ under the IBS policy. For a full derivation of the properties of the IBS estimator, we refer to Appendix \ref{sec:ibsworks}. \eq~\ref{eq:Libs} can be written compactly as $\Libs(K) = \psi(1) - \psi(K)$, where $\psi(z)$ is the \emph{digamma function} \citep{abramowitz1948handbook}.

We now provide an understanding of why fixed sampling is not a good policy, despite its intuitive appeal, and then show why IBS solves many of the problems with fixed sampling.

\section{Why fixed sampling fails}
\label{sec:fixed_fail}

\begin{figure}[htp]
  \centering
  \includegraphics[width=5.2in]{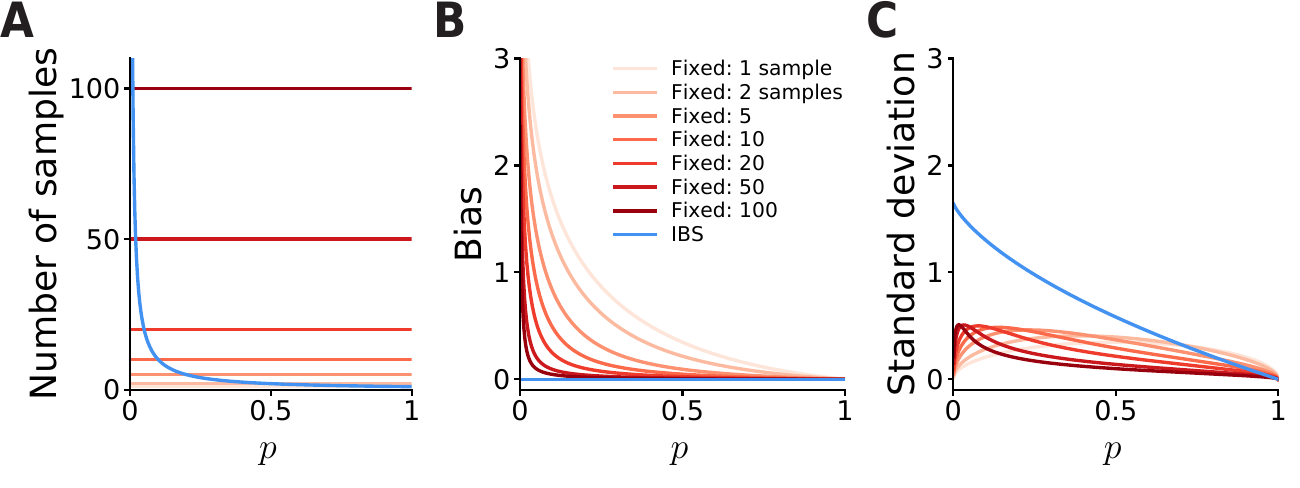}
  \vspace{-0.5em}
\caption{\textbf{A.} Number of samples used by fixed (red curves) or inverse binomial sampling (blue; expected value) to estimate the log-likelihood $\log p$ on a single trial with probability $p$. IBS uses on average $\frac{1}{p}$ trials. \textbf{B.} Bias of the log-likelihood estimate. The bias of IBS is identically zero. \textbf{C.} Standard deviation of the log-likelihood estimate.}
\label{fig:bias_variance_time}
\end{figure}

We summarize in Figure~\ref{fig:bias_variance_time} the properties of the IBS estimator and of fixed sampling, for different number of samples $M$, as a function of the trial likelihood $p$. In particular, we plot the expected number of samples, the bias, and the standard deviation of the estimators.


\begin{figure}[htp]
  \centering
  \includegraphics[width=5.2in]{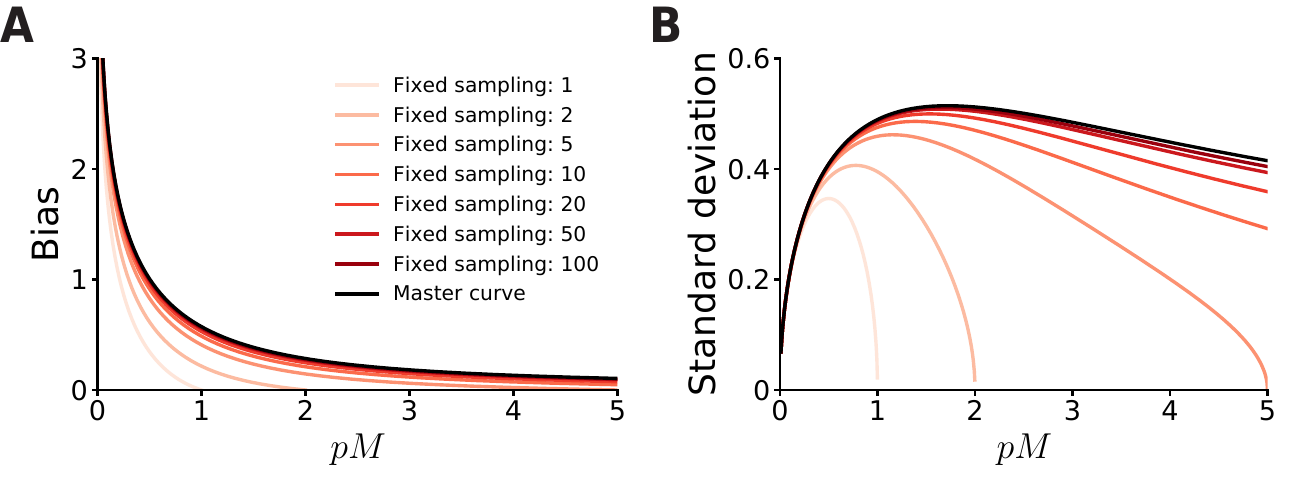}
  \vspace{-0.5em}
\caption{\textbf{A.} Bias of fixed sampling estimators of the log-likelihood, plotted as a function of $pM$, where $p$ is the likelihood on a given trial, and $M$ the number of samples. As $M\rightarrow\infty$, the bias converges to a master curve (\eq~\ref{eq:bias_master}). \textbf{B.} Same, but for standard deviation of the estimate.}
\label{fig:bias_variance_master}
\end{figure}

The critical disadvantage of the fixed sampling policy with $M$ samples is that its estimates of the log-likelihood are inevitably biased (see Figure~\ref{fig:bias_variance_time}B).
Fixed sampling is `inevitably' biased because the bias decreases as one takes more samples, but for $p\rightarrow 0$, the estimator remains biased. More precisely, in a joint limit where $M\rightarrow\infty$, $p\rightarrow 0$ and $pM\rightarrow \lambda$ for some constant $\lambda$, the bias collapses onto a single `master curve' (see Figure~\ref{fig:bias_variance_master}; and Appendix \ref{sec:analysisbias} for the derivation). In particular, we observe that the bias is close to zero for $\lambda\gg 1$ and that it diverges when $\lambda\ll 1$, or equivalently, for $M\gg\frac{1}{p}$ and $M\ll\frac{1}{p}$, respectively.

To convey the intuition for why the bias diverges for small probabilities, we provide a gambling analogy. Imagine playing a slot machine and losing the first $100$ bets you make. You can now deduce that this slot machine likely has a win rate less than $1\%$, but there is no way of knowing whether it is $1\%$, $0.1\%$, $0.01\%$ or even $0\%$ apart from any prior beliefs you may have (for example, you expect that the house has stacked the odds in their favor but not overwhelmingly so).
In practice, this uncertainty is unlikely to affect your decision whether to continue playing the slot machine, since the expected value of the slot machine depends linearly on its win rate. However, if your goal is to estimate the \emph{logarithm} of the win rate, the difference between these percentages becomes infinitely large as the true win rate tends to 0. 
We provide a more formal treatment of the bias of fixed sampling in Appendix \ref{sec:analysisbias}.

\subsection{Why fixed sampling cannot be fixed}
\label{sec:unfixable}

The asymptotic analyses above suggest an obvious solution to prevent fixed sampling estimators from becoming strongly biased: make sure to draw enough samples so that $M\gg\max_{i=1\dots N}\frac{1}{p_i}$. Although this solution will succeed in theory, it has practical issues. First of all, choosing $M$ requires knowledge of $p_i$ on each trial, which is equivalent to the problem we set out the solve in the first place. Moreover, even if one can derive or estimate an upper bound on $\frac{1}{p_i}$ (for example, in behavioral models that include a lapse rate, that is a nonzero probability of giving a uniformly random response), fixed sampling will be inefficient. As shown in Figure~\ref{fig:bias_variance_master}, the bias in $\hat{\L}_{\text{fixed}}$ is small when $\lambda\approx 1$ or $M\approx\frac{1}{p}$ and increasing $M$ even further has diminishing returns, at least for the purpose of reducing bias. If we choose $M$ inversely proportional to the probability $p_i$ on the trial where the model is least likely to match the observed response, we will draw many more samples than necessary for all other trials. 

One might hope that in practice the likelihood $p_i$ is approximately the same across trials, but the opposite is true. As an example, take a typical `orientation discrimination' psychophysical task in which a participant has to detect whether a presented oriented grating is tilted clockwise or anti-clockwise from vertical, and consider a generative model for the observer's responses that includes sensory measurement noise and lapses (see Section~\ref{sec:ori} for details).
Moreover, imagine that the experiment contains $\approx 500$ trials, and the participant's true lapse rate is $1\%$. The model will always assign more probability to correct responses than errors, so, for all correct trials, $p_i$ will be at least $0.5$. However, there will likely be a handful of trials where the participant lapses and makes a grave error (responding incorrectly to a stimulus very far from the decision boundary), in which case $p_i$ will be $0.5$ times the lapse rate. This hypothetical scenario is not exceptional, in fact it is almost inevitable in any experiment where participants occasionally make unpredictable responses, and perform hundreds or more trials. 

A more sophisticated solution would relax the assumption that $M$ needs to be constant for all trials, and instead choose $M$ as a function of $p_i$ on each trial. However, since $p_i$ is unknown, one would need to first estimate $p_i$ by sampling, choose $M_i$ for each trial, then re-estimate $\L_i$. Such an iterative procedure would create a non-fixed sampling scheme, in which $M_i$ adapts to $p_i$ on each trial. This approach is promising, and it is, in fact, how we originally arrived at the idea of using inverse binomial sampling for log-likelihood estimation, while working on the complex cognitive model described in Section \ref{sec:fourinarow}.

Finally, a heuristic solution would be to disregard any statistical concerns, pick $M$ based on some intuition or from similar studies in the literature, and hope that the bias turns out to be negligible. We do not intend to dissuade researchers from using such pragmatic approaches if they work in practice. Unfortunately, this one does not. As Figure~\ref{fig:bias_variance_master} shows, estimating log-likelihoods with fixed sampling can cause biases of $1$ or more points of model evidence if the data set contains even a single trial on which $p_i\leq\frac{1}{2M}$. Since differences in log-likelihoods larger than $5$ to $10$ points are often regarded as strong evidence for one model over another~\citep{kass1995bayes,jeffreys1998theory,anderson2002model}, it is well possible for such biases to reverse the outcome of a model comparison analysis. 
This point bears repeating; if one uses fixed sampling to estimate log-likelihoods and the number of samples is too low, one risks of drawing conclusions about scientific hypotheses that are not supported by the experimental data one has collected. 

\subsection{Why not an unbiased estimator of the likelihood?}
\label{sec:whynotlikelihood}

In this paper, we focus on finding an unbiased estimator of the log-likelihood, but one might wonder why we do not look instead for an unbiased estimator of the likelihood. In fact, we already have such an estimator. Fixed sampling provides an unbiased estimate of $p_i$ for each trial, and since all estimates are unbiased and statistically independent, $\prod_i p_i$ is also unbiased. 

The critical issue is the shape of the distribution of these estimates. While a central limit theorem for products of random variables exists, it only holds if all the estimates are almost surely not zero. For estimates of $p_i$ obtained via fixed sampling, this is not the case, and we would not obtain a well-behaved (i.e., log-normal) distribution of $\prod_i p_i$. In fact, the distribution would be highly multimodal with the main peak being at $\prod_i p_i = 0$. This property makes this estimator unusable for all practical purposes (e.g., from maximum-likelihood estimation to Bayesian inference).

Instead, by switching to log-likelihood estimation, we can find an estimator (IBS) which is both unbiased and whose estimates are guaranteed to be well-behaved (in particular, normally distributed). 
We stress that normality is not just a desirable addition, but a fundamental feature with substantial practical consequences for how estimators are used, as we will see more in detail in the following section.



\section{Is inverse binomial sampling really better?}
\label{sec:ibs_better}

While one could expect that the unbiasedness of the IBS estimator would come at a cost, such as more samples, a much higher variance, or perhaps a particularly complex implementation, we show here that IBS is not only unbiased, but it is sample-efficient, its estimates are low-variance, and can be implemented in a few lines of code.

\subsection{Implementation}
\label{sec:implementation}

We present in Algorithm \ref{alg:ibs_basic} a description in pseudo-code of the basic IBS procedure to estimate the log-likelihood of a given parameter vector $\vtheta$ for a given data set and generative model. 
The procedure is based on the inverse binomial sampling scheme introduced in Section~\ref{sec:policies}, generalized sequentially to multiple trials.

For each trial, we draw sampled responses from the generative model, given the stimulus $\s_i $ in that trial, using the subroutine \texttt{sample\_from\_model}, until one matches the observed response $\r_i$. This yields a value of $K_i$ on each trial $i$, which IBS converts to an estimate $\hat{\L}_i$ (where we use the convention that a sum with zero terms equals 0). We make our way sequentially across all trials, returning then the summed log-likelihood estimate $\Libs$ for the entire data set.

\begin{algorithm}[ht]
\caption{Inverse Binomial Sampling (sequential implementation)}\label{alg:ibs_basic}
\begin{algorithmic}[1]
\INPUT Stimuli $\left\{\s_i\right\}_{i = 1}^N$, 
responses $\left\{\r_i\right\}_{i=1}^N$, 
generative model $\mathcal{M}$, parameters $\vtheta$
\For{$i \leftarrow 1 \ldots N$} \Comment{Sequential loop over all trials}
\State $K_i \leftarrow 1$
\While{\texttt{sample\_from\_model}($\mathcal{M}$,$\vtheta$,$\s_i$) $\neq \r_i$}
\State $K_i \leftarrow K_i + 1$
\EndWhile
\State $\hat{\L}_i \leftarrow - \displaystyle\sum_{k=1}^{K_i-1} \frac{1}{k}$ \Comment IBS estimator from \eq~\ref{eq:Libs}
\EndFor
\State \Return $\displaystyle\sum_{i=1}^N \hat{\L}_i$  \Comment{Return total log-likelihood estimate}
\end{algorithmic}
\end{algorithm}


In practice, depending on the programming language of choice, it might be useful to take advantage of numerical features such as vectorization to speed up computations.  An alternative `parallel' implementation of IBS is described in Appendix \ref{sec:threshold}. 

One might wonder how to choose the $\vtheta$ to evaluate in Algorithm \ref{alg:ibs_basic} in the first place. IBS is agnostic of how candidate $\vtheta$ are proposed. Most often, the function that implements Algorithm \ref{alg:ibs_basic} will be passed to a chosen optimization or inference algorithm, and the job of proposing $\vtheta$ will be taken care of by the chosen method. For maximum-likelihood estimation, we recommend derivative-free optimization methods that deal effectively and robustly with noisy evaluations, such as Bayesian Adaptive Direct Search (BADS; \citealp{acerbi2017practical}) or noise-robust CMA-ES \citep{hansen2003reducing,hansen2008method}.
At the end of optimization, most methods will then return a candidate solution $\widehat{\vtheta}_\text{MLE}$ and possibly an estimate of the value of the target function at $\widehat{\vtheta}_\text{MLE}$. However, since the target function is noisy and the final estimate is biased (because, by definition, it is better than the other evaluated locations), we recommend to re-estimate $\Libs(\widehat{\vtheta}_\text{MLE})$ multiple times to obtain a higher-precision, unbiased estimate of the log-likelihood at $\widehat{\vtheta}_\text{MLE}$ (see also Section \ref{sec:multifidelity}).

Implementations of IBS in different programming languages can be found at the following web page: \url{https://github.com/lacerbi/ibs}.

\subsection{Computational time}
\label{sec:time}

The number of samples that IBS takes on a trial with probability $p_i$ is geometrically distributed with mean $\frac{1}{p_i}$. We saw earlier that for fixed-sampling estimators to be approximately unbiased, one needs at least $\frac{1}{p_i}$ samples, and IBS does exactly that in expectation.
Moreover, since IBS adapts the number of samples it takes on different trials, it will be considerably more sample-efficient than fixed sampling with constant $M$ across trials. For example, in the aforementioned example of the orientation discrimination task, when most trials have a likelihood $p_i\geq 0.5$, IBS will often take just $1$ or $2$ samples on those trials. Therefore, it will allocate most of its samples and computational time on trials where $p_i$ is low and those samples are needed.

\subsection{Variance}
\label{sec:variance}

The variance of the IBS estimator can be derived as
\begin{equation} \label{eq:ibsvariance}
\text{Var}\left[\Libs\right]=\sum_{k=1}^{\infty}\frac{1}{k^2}(1-p)^k=\text{Li}_2(1-p),
\end{equation}
where we introduced the dilogarithm or Spence's function $\text{Li}_2(z)$ \citep{maximon2003dilogarithm}. The variance (plotted in Figure~\ref{fig:bias_variance_time}C as standard deviation) increases when $p\rightarrow 0$, but it does not diverge; instead, it converges to $\frac{\pi^2}{6}$. 
Therefore, IBS is not only uniformly unbiased, but its variance is uniformly bounded. The full derivation of Equation \ref{eq:ibsvariance} is reported in Appendix \ref{sec:derivation_ibs_var}.

We already mentioned that $\Libs$ is the minimum-variance unbiased estimator of $\log p$ given the inverse binomial sampling policy, but it also comes \emph{close} (less than $\sim30 \%$ distance) to saturating the \emph{information inequality}, which specifies the minimum variance that can be theoretically achieved by any estimator under a non-fixed sampling policy (an analogue of the Cramer-R\'ao bound; \citealp{degroot1959unbiased}). We note that fixed sampling eventually saturates the information inequality in the limit $M \rightarrow \infty$, but as mentioned in the previous section, the fixed-sampling approach can be highly wasteful or substantially biased (or both), not knowing a priori how large $M$ has to be across trials. See Appendix \ref{sec:infoinequality} for a full discussion of the information inequality and comparison between estimators.

\eq~\ref{eq:ibsvariance} has theoretical relevance, but requires us to know the true value of the likelihood $p$, which is unknown in practice. Instead, we define the estimator of the variance of a specific IBS estimate, having sampled for $K$ times until a `hit', as
\begin{equation} \label{eq:ibsvarest}
\text{Var}\left[\left.\Libs \right| K \right] = \psi_1(1) - \psi_1(K),
\end{equation}
where $\psi_1(z)$ is the \emph{trigamma function}, the derivative of the digamma function \citep{abramowitz1948handbook}. We derived \eq~\ref{eq:ibsvarest} from a Bayesian interpretation of the IBS estimator, which can be found in Appendix \ref{sec:ibs_bayesian}.
Note that Equations \ref{eq:ibsvariance} and \ref{eq:ibsvarest} correspond to slightly different concepts, in that the former represents the variance of the estimator for a known $p$ (from a frequentist point of view), while the latter is the posterior variance of $\L$ for a known $K$ (for which there is no frequentist analogue). See also Section \ref{sec:highermoments} for further discussion.

\subsection{Iterative multi-fidelity}
\label{sec:multifidelity}

We define a \emph{multi-fidelity} estimator as an estimator with a tunable parameter that affords different degrees of precision at different computational costs (i.e., from a cheaper, inaccurate estimate to a very accurate but expensive one), borrowing the term from the literature on computer simulations and surrogate models \citep{kennedy2000predicting,forrester2007multi}. IBS provides a particularly convenient way to construct an \emph{iterative} multi-fidelity estimator in that we can perform $\Reps$ independent `repeats' of the IBS estimate at $\vtheta$, and combine them by averaging,
\begin{equation} \label{eq:repeats}
\begin{split}
\Libsrep{\Reps}(\vtheta) = & \frac{1}{\Reps} \sum_{\reps = 1}^\Reps \Libs^{(\reps)}(\vtheta) \\
\text{Var}\left[\Libsrep{\Reps}(\vtheta)\right] = & \frac{1}{\Reps^2} \sum_{\reps=1}^\Reps \text{Var}\left[\Libs^{(\reps)}(\vtheta)\right], \\
\end{split}
\end{equation}
where $\Libs^{(\reps)}$ denotes the $\reps$-th independent estimate obtained via IBS. For $\Reps=1$, we recover the standard (`1-rep') IBS estimator.  The variances in \eq~\ref{eq:repeats} are computed empirically using the estimator in \eq~\ref{eq:ibsvarest}.

Importantly, we do not need to perform all $\Reps$ repeats at the same time, but we can iteratively refine our estimates whenever needed, and only need to store the current estimate, its variance and the number of repeats performed so far:
\begin{equation}
\begin{split}
\Libsrep{\Reps+1}(\vtheta) = & \frac{1}{\Reps+1} \left[ \Reps \cdot \Libsrep{\Reps}(\vtheta) + \Libs^{(\reps+1)}(\vtheta) \right]\\
\text{Var}\left[\Libsrep{\Reps+1}(\vtheta)\right] = & \frac{1}{(\Reps+1)^2} \left\{ \Reps^2 \cdot \text{Var}\left[\Libsrep{\Reps}(\vtheta)\right] + \text{Var}\left[\Libs^{(\reps+1)}(\vtheta)\right] \right\}.
\end{split}
\end{equation}
Crucially, while a similar procedure could be performed with any estimator (including fixed sampling), the fact that IBS is unbiased and its variance is bounded ensures that the combined iterative estimator is also unbiased and eventually converges to the true value for $\Reps \rightarrow \infty$, with variance bounded above by $\frac{\pi^2}{6 \Reps}$.

Finally, we note that the iterative multi-fidelity approach described in this section can be extended such that, instead of having the same number of repeats $R$ for all trials, one could adaptively allocate a different number of repeats $R_i$ to each trial so as to minimize the overall variance of the estimated log-likelihood (see Appendix \ref{sec:repeated_sampling}).

\subsection{Bias or variance?}

In the previous sections, we have seen that IBS is always unbiased, whereas fixed sampling can be highly biased when using too few samples. However, with the right choice of $M$, fixed sampling can have lower variance. We now list several practical and theoretical arguments for why bias can have a larger negative impact than variance, and being unbiased is a desirable property for an estimator of the log-likelihood.
\begin{enumerate}
\item To use IBS or fixed sampling to estimate the log-likelihood of a given data set, we sum estimates of $\L_i$ across trials. Being the sum of independent random variables, as $N\rightarrow\infty$, the standard deviation of $\hat{\L}\left(\vtheta\right)$ will grow proportional to $\sqrt{N}$, whereas the bias grows linearly with $N$. For a concrete example, see Appendix \ref{sec:estimator_rmse}.
\item When using the log-likelihood (or a derived metric) for model selection, it is common to collect evidence for a model, possibly hierarchically, across multiple datasets (e.g., different participants in a behavioral experiment), which provides a second level of averaging that can reduce variance but not bias. 
\item Besides model selection, the other key reason to estimate log-likelihoods is to infer parameters of a model, for example via maximum-likelihood estimation. For this purpose, one would use an optimization algorithm that calls the routine that estimates $\hat{\L}\left(\vtheta\right)$ many times with different candidate values of $\vtheta$, and uses this information to estimate the value that maximizes $\L\left(\vtheta\right)$. Powerful, sample-efficient optimization algorithms, such as those based on Bayesian optimization, work by building a statistical approximation (a \emph{surrogate}) of the objective function \citep{jones1998efficient,snoek2012practical,shahriari2015taking,acerbi2017practical}, most commonly via Gaussian processes \citep{rasmussen2006gaussian}. These methods can operate successfully with noisy objectives by effectively averaging function values from nearby parameter vectors. By contrast, no optimization algorithm can handle bias. 
This argument is not limited to maximum-likelihood estimation, as recent methods have been proposed to use Gaussian process surrogates to perform (approximate) Bayesian inference and infer posterior distributions \citep{kandasamy2015bayesian,acerbi2018variational,acerbi2020variational,jarvenpaa2019parallel}; also these techniques can handle variance in the estimates but not bias.
\item The ability to combine unbiased estimates of known variance iteratively (as described in Section \ref{sec:multifidelity}) is particularly useful with adaptive fitting methods based on Gaussian processes, whose algorithmic cost grows super-linearly in the number of \emph{distinct} training points \citep{rasmussen2006gaussian}. Thanks to iterative multi-fidelity estimation, these methods would have the opportunity to refine their estimates of the log-likelihood at a previously evaluated point, whenever deemed useful, without incurring an increased algorithmic cost.
\item On a conceptual level, bias is potentially more dangerous than variance. Bias can cause researchers to confidently draw false conclusions, variance, when properly accounted for, causes decreased statistical power and lack of confidence. Appropriate statistical tools can account for variance and explain seemingly conflicting findings resulting from underpowered studies \citep{maxwell2015psychology}, whereas bias is much harder to recognize or correct no matter what statistical techniques one uses.
\end{enumerate}
Finally, while for the sake of argument we structured this section as an opposition between bias and variance, it is not an exact dichotomy and both properties matter, together with other considerations (see Appendix \ref{sec:estimator_rmse}). For example, there are situations in which the researcher may knowingly decide to increase bias to reduce both variance and computational costs. Notably, while this trade-off is easy to achieve with the IBS estimator (see Appendix \ref{sec:threshold}), there is no similarly easy technique to de-bias a biased estimator.


\subsection{Higher-order moments}
\label{sec:highermoments}

So far, we have considered the mean (or bias) and variance of $\hat{\L}_{\text{fixed}}$ and $\Libs$ in detail, but ignored any higher-order moments. This is justified since to estimate the log-likelihood of a model with a given parameter vector we will sum these estimates across many trials. Therefore, the central limit theorem guarantees that the distribution of $\L\left(\vtheta\right)$ is Gaussian with mean and variance determined by the mean and variance of $\hat{\L}_{\text{fixed}}$ or $\Libs$ on each trial, at least as long as the distribution of $p_i$ across trials satisfies a regularity condition.\footnote{Specifically, the \citet{lindeberg1922neue} or Lyapunov conditions \citep[Chapter 7.3]{ash2000probability}, both of which place restrictions on the degree to which the variance of any single trial can dominate the distribution of $\displaystyle\sum_i\hat{\L}_i$.} A sufficient but far from necessary condition is that there exists a lower bound on $p_i$, which is the case for example for a behavioral model with a lapse rate. Using the same argument, the total number of samples $K_\text{tot}$ that IBS uses to estimate $\L\left(\theta\right)$ is also approximately Gaussian. 


\begin{figure}[htp]
  \centering
  \includegraphics[width=5.2in]{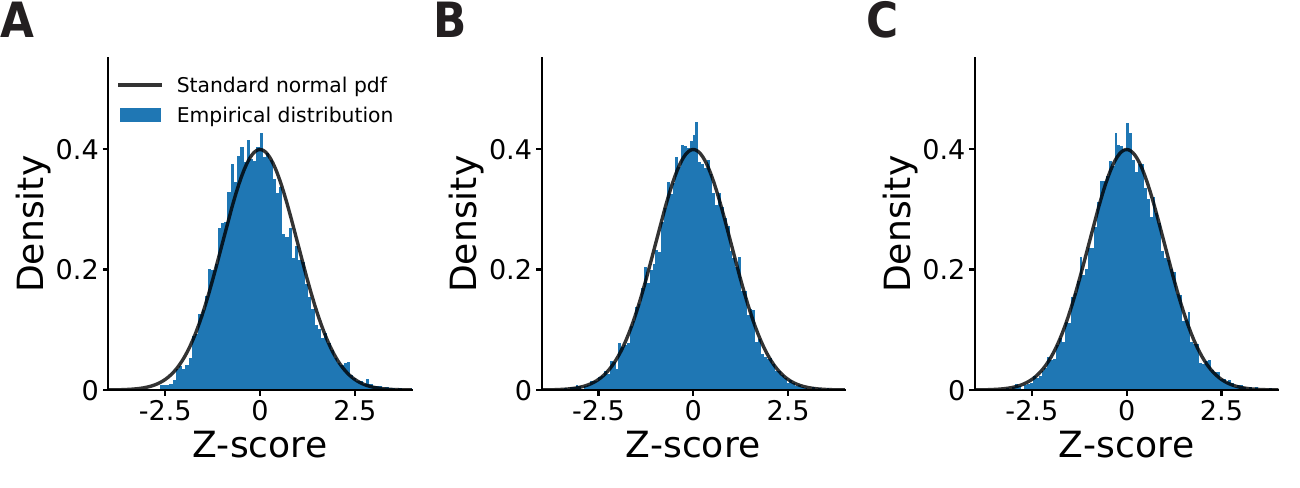}
  \vspace{-0.5em}
\caption{\textbf{A.} $z$-score plot for the total number of samples used by IBS. \textbf{B.} $z$-score plot for the estimates returned by IBS, using the exact variance formula for known probability. \textbf{C.} Calibration plot for the estimates returned by IBS, using the variance estimate from \eq~\ref{eq:ibsvarest}. These figures show that the number of samples taken by IBS and the estimated log-likelihood are (approximately) Gaussian, and that the variance estimate from \eq~\ref{eq:ibsvarest} is calibrated.}
\label{fig:calibration}
\end{figure}

In the following, we demonstrate empirically that the distributions of the number of samples taken by IBS and of the estimates $\Libs$ are Gaussian. Importantly, we also show that the estimate of the variance from \eq~\ref{eq:ibsvarest}, $\hat{V}_\text{IBS}$, is \emph{calibrated}. That is, we expect the fraction of estimates within the credible interval $\Libs \pm \beta \sqrt{\hat{V}_\text{IBS}}$ to be (approximately) $\Phi(\beta) - \Phi(-\beta)$, where $\Phi(x)$ is the cumulative normal distribution function and $\beta > 0$.

As a realistic scenario, we consider the psychometric function model described in Section \ref{sec:ori}. For each simulated data set, we estimated the log-likelihood under the true data-generating parameters $\vtheta_\text{true}$ (see Appendix \ref{sec:ori_details} for details). Specifically, for each data set we ran IBS and recorded the estimated log-likelihood $\Libs$, the total number of samples $K_\text{tot}$ taken, and a Bayesian estimate for the variance of $\Libs$ from \eq~\ref{eq:ibsvarest}. For the total number of samples $K_\text{tot}$ and the $\Libs$ estimate, we can compute the theoretical mean and variance by knowing the trial likelihoods $p_i$, which we can evaluate exactly in this example. 

For each obtained $K_\text{tot}$, we computed a $z$-score by subtracting the exact mean and dividing by the exact standard deviation, obtained by knowing the mean and variance of geometric random variables underlying the samples taken in each trial. If $K_\text{tot}$ is normally distributed, we expect that the variable $z$ across data sets should appear to be distributed as a standard normal, $z \sim \mathcal{N}\left(0,1\right)$. If $K_\text{tot}$ is not normally distributed, we should see deviations from normality in the distribution of $z$, especially in the tails. By comparing the histogram of $z$-scores with a standard normal in Figure \ref{fig:calibration}A, we see that the total number of samples is approximately normal, with some residual skew.

We did the same analysis for the estimate $\Libs$, using the $z$-scored variable
\begin{equation}\label{eq:zscore}
z \equiv \frac{\Libs - \L_\text{true}}{\sqrt{\text{Var}[\Libs]}},
\end{equation}
where here $\text{Var}[\Libs]$ is the exact variance of the estimator computed via \eq~\ref{eq:ibsvariance}. The histogram of $z$-scores in Figure~\ref{fig:calibration}B is again very close to a standard normal.

Finally, in practical scenarios we do not know the true likelihoods, so the key question is whether we can obtain valid estimates of the variance of $\Libs$ via \eq~\ref{eq:ibsvarest}. If such an estimate is correctly calibrated, the distribution of $z$-scores should remain approximately Gaussian if we use \eq~\ref{eq:ibsvarest} for the denominator of \eq~\ref{eq:zscore}. Indeed, the calibration plot in Figure~\ref{fig:calibration}C shows an excellent match with a standard normal, confirming that our proposed estimator of the variance is well calibrated.

\section{Numerical experiments}
\label{sec:experiments}

In this section, we examine the performance of IBS and fixed sampling on several realistic model-fitting problems of increasing complexity.
The example problems we consider here model tasks drawn from psychophysics and cognitive science: an orientation discrimination experiment (Section \ref{sec:ori}); a change localization task (Section \ref{sec:change}); and playing a four-in-a-row game that involves complex sequential decision making (Section \ref{sec:fourinarow}). For the first problem, we can derive the exact analytical expression for the log-likelihood; for the second problem, we have an integral expression for the log-likelihood that we can approximate numerically; and finally, for the third problem, we are in the true scenario in which the log-likelihood is intractable.

The rationale for our numerical experiments is that so far we have analyzed fixed sampling and IBS in terms of the bias in their log-likelihood estimates for individual parameter vectors. However, these log-likelihood estimators are often used as elements of a more complex statistical procedure, such as maximum-likelihood estimation. It is plausible that biases in log-likelihood estimates will lead to biases in parameter estimates obtained by maximizing the log-likelihood, but the exact relationship between those biases, and the role of variance in optimization is not immediate. Similarly, it is unclear how bias and variance of individual log-likelihood estimates will affect the estimate of the \emph{maximum} log-likelihood, often used for model selection (e.g., unbiased estimates of the log-likelihood do not guarantee an error-free estimate of the maximum log-likelihood, which is affected by other factors; see Section \ref{sec:loglikeloss}).
Therefore, we conduct an empirical study showing that, in practice, IBS leads to more accurate parameter and maximum log-likelihood estimates than fixed sampling, given the same budget of computational resources.

First, we describe the procedure used to perform our numerical experiments. Code to run all our numerical experiments and analyses is available at the following link: \texttt{https://github.com/basvanopheusden/ibs-development}.

\subsection{Procedure}
\label{sec:procedure}

For each problem, we simulate data from the generative model given different known settings $\vtheta_\text{true}$ of model parameters, and we compare the accuracy (and other statistics) of both IBS and fixed sampling in recovering the true data-generating parameters through maximum-likelihood estimation.
Since these methods provide noisy and possibly biased estimates of $\L\left(\vtheta\right)$, and due to variability in the simulated datasets, the estimates $\widehat{\vtheta}_\text{MLE}$ that result from optimizing the log-likelihood will also be noisy and possibly biased.
To explore performance in a variety of settings, and to account for variability in the data-generation process, for each problem we consider $40 \cdot D$ different parameter settings, where $D$ is the number of model parameters (that is, the dimension of $\vtheta$), and for each parameter setting we generate $100$ distinct synthetic datasets.

For each dataset, we compare fixed sampling with different numbers of samples $M$ (from $M = 1$ to $M = 50$ or $M = 100$, depending on the problem), and IBS with different number of `repeats' $R$, as defined in Section \ref{sec:multifidelity} (from $R=1$ to up to $R = 50$, depending on the problem). 
In each scenario, we directly compare the two methods in terms of number of samples by computing the \emph{average} number of samples used by IBS for a given number of repeats $R$.
To prevent IBS from `hanging' on particularly bad parameter vectors, we adopt the `early stopping threshold' technique described in Appendix \ref{sec:threshold}.
Finally, if available, we also test the performance of maximum-likelihood estimation using the `exact' log-likelihood function (calculated either analytically or via numerical integration).

For all methods, we maximize the log-likelihood with Bayesian Adaptive Direct Search (BADS, \citealp{acerbi2017practical}; \texttt{github.com/lacerbi/bads}), a hybrid Bayesian optimization algorithm based on the mesh-adaptive direct search framework \citep{audet2006mesh}, which affords a fast, robust exploration of the function landscape via Gaussian process surrogates. Briefly, BADS works by alternating between two stages: in the Poll stage, the algorithm evaluates points in a random mesh surrounding the current point, in a fairly model-free way; in the Search stage, following the principles of Bayesian optimization \citep{jones1998efficient}, the algorithm builds a local Gaussian process model of the target function, and chooses the next point by taking into account both mean and variance of the surrogate model, balancing exploration of unknown but promising regions and exploitation of regions known to be high-valued (for maximization).
By combining model-free and powerful model-based search, BADS has been shown to be much more effective than alternative optimization methods particularly when dealing with stochastic objective functions, and with a relatively limited budget of a few hundreds to a few thousands function evaluations \citep{acerbi2017practical}.
We refer the interested reader to \citet{acerbi2017practical} and to the extensive online documentation for further information about the algorithm.

\subsection{Orientation discrimination}
\label{sec:ori}

\begin{figure}[htp]
  \centering
  \includegraphics[width=5.2in]{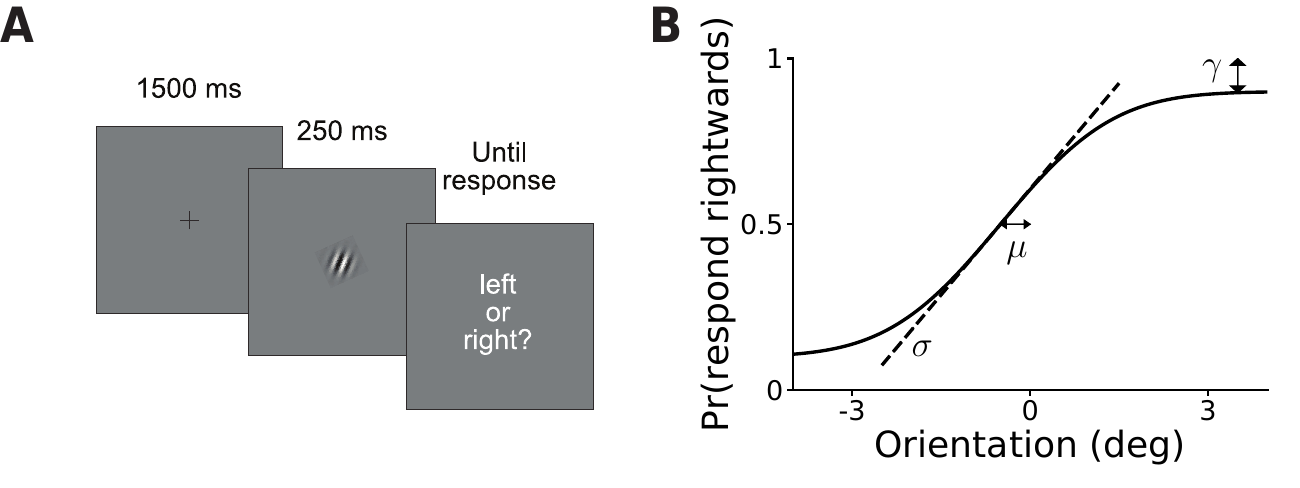}
  \vspace{-0.5em}
\caption{\textbf{A.} Trial structure of the simulated orientation discrimination task. A oriented patch appears on a screen for $250$ ms, after which participants decide whether it is rotated rightwards or leftwards with respect to a vertical reference. 
\textbf{B.} Graphical illustration of the behavioral model, which specifies the probability of choosing rightwards as a function of the true stimulus orientation. The three model parameters $\sigma$, $\mu$, and $\lapse$ correspond to the (inverse) slope, horizontal offset and (double) asymptote of the psychometric curve, as per Equation \ref{eq:ori_equation}. Note that we parametrize the model with $\eta \equiv \log\sigma$.}
\label{fig:ori_task_model}
\end{figure}

The first task we simulate is an orientation discrimination task, in which a participant observes an oriented patch on a screen, and indicates whether they believe it was rotated leftwards or rightwards with respect to a reference line (see Figure~\ref{fig:ori_task_model}A). Here, on each trial the stimulus $s$ is the orientation of the patch with respect to the reference (in degrees), and the response $r$ is `rightwards' or `leftwards'.

For each dataset, we simulated $N = 600$ trials, drawing on each trial the stimulus $s$ from a Gaussian distribution with mean $0^{\circ}$ (the reference) and standard deviation $3^{\circ}$. 
The generative model assumes that the observer makes a noisy measurement $x$ of the stimulus, which is normally distributed with mean $s$ and standard deviation $\sigma$, as per standard signal detection theory \citep{green1966signal}. They then respond `rightwards' if $x$ is larger than $\mu$ (a parameter which captures response bias, or an incorrect memory of the reference line) and `leftward' otherwise. However, a fraction of the time, given by the lapse rate $\gamma \in (0, 1]$, the observer guesses randomly. We visually illustrate the model in Figure~\ref{fig:ori_task_model}B. For both theoretical reasons and numerical convenience, we parametrize the slope $\sigma$ as $\eta \equiv \log \sigma$. Thus, the model has parameter vector $\vtheta = (\eta, \mu, \lapse)$.  

We can derive the likelihood of each trial analytically: 
\begin{equation} \label{eq:ori_equation}
\Pr(\text{`rightwards' response}\lvert s,\vtheta)=\frac{\lapse}{2}+(1-\lapse)\Phi\left(\frac{s-\mu}{\sigma}\right),
\end{equation}
where $\Phi(x)$ is the cumulative normal distribution function. \eq~\ref{eq:ori_equation} takes the form of a typical psychometric function ~\citep{wichmann2001psychometric}. 
Note that in this section we use Gaussian distributions for circularly distributed variables, which is justified under the assumption that both the stimulus distribution and the measurement noise are small. For more details about the numerical experiments, see Appendix~\ref{sec:ori_details}.

\begin{figure}[htp]
  \centering
  \includegraphics[width=5.2in]{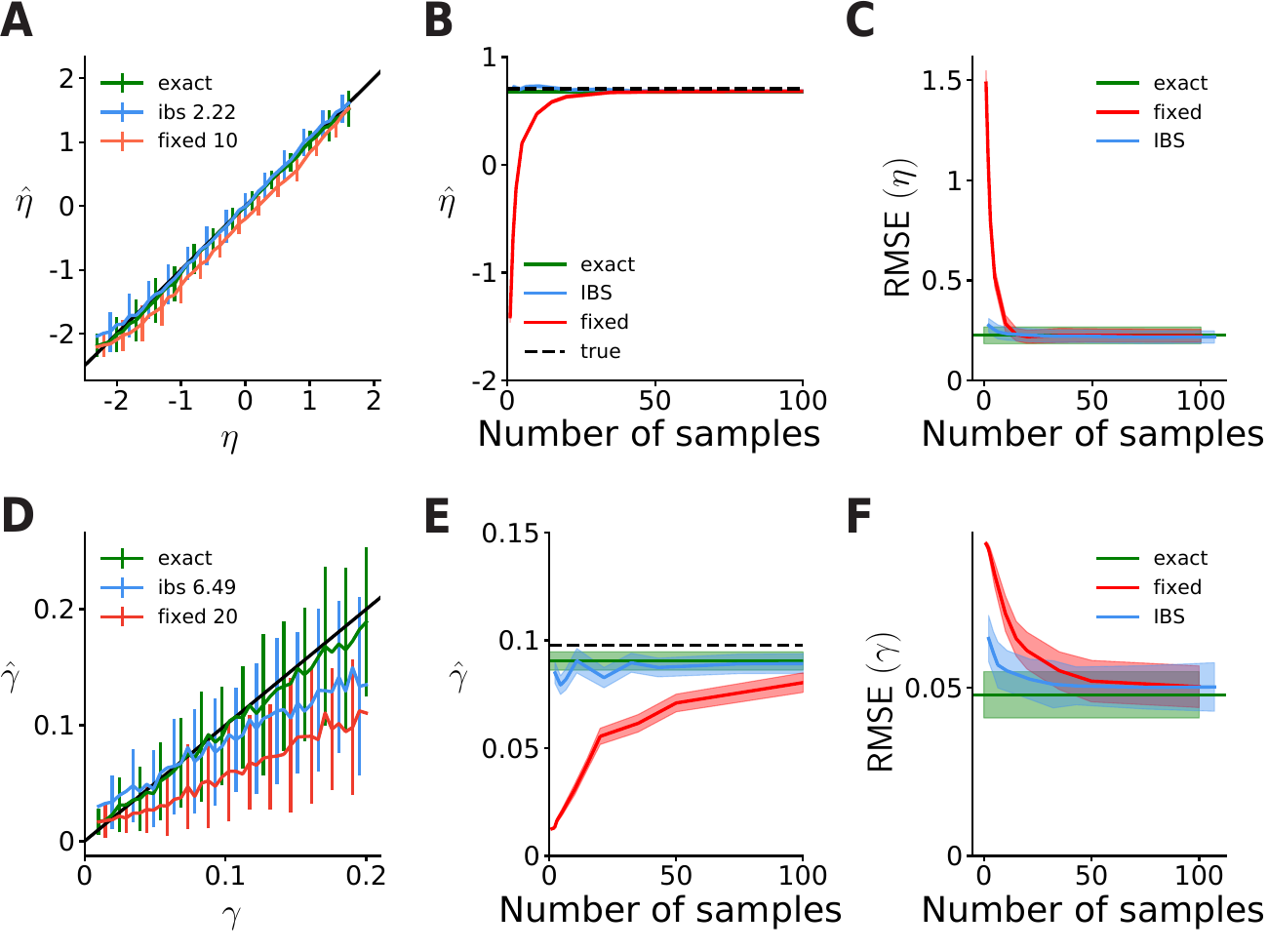}
  \vspace{-0.5em}
\caption{ 
\textbf{A.} Estimated values of $\eta \equiv \log \sigma$ as a function of the true $\eta$ in simulated data using IBS with $R=1$ repeat (blue), fixed sampling with $M=10$ (red) or the exact likelihood function (green). The black line denotes equality. Error bars indicate standard deviation across $100$ simulated data sets. IBS uses on average $2.22$ samples per trial. \textbf{B.} Mean and standard error (shaded regions) of estimates of $\eta$ for $100$ simulated data sets with $\eta_{\text{true}}= \log 2^{\circ}$, using fixed sampling, IBS or the exact likelihood function. For fixed sampling and IBS, we plot mean and standard error as a function of the (average) number of samples used. \textbf{C.} Root mean squared error (RMSE) of estimates of $\eta$, averaged across the range of $\eta_{\text{true}}$ in \textbf{A}, as a function of the number of samples used by IBS or fixed sampling. Shaded regions denote $\pm 1$ standard deviation across the $100$ simulated data sets. We also plot the RMSE of exact maximum-likelihood estimation, which is nonzero since we simulated data sets with only $600$ trials. \textbf{D-F} Same, for $\lapse$ (with $R = 3$, $M = 20$ in panel \textbf{D}). These results demonstrate that IBS estimates parameters of the model for orientation discrimination more accurately than fixed sampling using equally many or even fewer samples.
} 
\label{fig:ori_results}
\end{figure}

In Figure~\ref{fig:ori_results}, we show the parameter recovery using fixed sampling, IBS and the exact log-likelihood function from \eq~\ref{eq:ori_equation}. First, we show that IBS can estimate the sensory noise parameter $\eta$ and lapse rate $\lapse$ more accurately than fixed sampling while using on average the same or fewer samples (Figure~\ref{fig:ori_results}A,D). For visualization purposes, we show here a representative example with $R = 1$ or $R=3$ repeats of IBS and $M = 10$ or $M = 20$ fixed samples (see Figure \ref{fig:psycho_complete_results} in the Appendix for the plots with all tested values of $R$ and $M$). As baseline, we also plot the mean and standard deviation of exact maximum-likelihood estimation, which is imperfect due to the finite data size ($600$ trials), and stochasticity and heuristics used in the optimization algorithm. We omit results for estimates of the response bias $\mu$, since even fixed sampling can match the performance of exact MLE with only $1$ sample per trial.

Next, we fix $\eta_\text{true} \equiv \log \sigma_\text{true} = \log 2^{\circ}, \mu_\text{true}=0.1^{\circ},\lapse_\text{true} =0.1$ and plot the mean and standard deviation of the estimated $\hat{\eta}$ and $\hat{\lapse}$ across $100$ simulated data sets as a function of the (average) number of samples per trial used by IBS or fixed sampling (Figure~\ref{fig:ori_results}B,E). We find that fixed sampling is highly sensitive to the number of samples, and with less than $20$ samples per trial, its estimate of $\eta$ is strongly biased. Estimating $\lapse$ accurately remains unattainable even with $100$ samples per trial. By contrast, IBS estimates $\eta$ and $\lapse$ reasonably accurately regardless of the number of samples per trial. IBS has a slight tendency to underestimate $\lapse$, which is a result of an interaction of the uncertainty handling in BADS with our choice of model parametrization and parameter bounds. In general, estimating lapse rates is notoriously prone to biases~\citep{prins2012psychometric}.

Finally, we measure the root mean squared error (RMSE) of IBS, fixed sampling and the exact solution, averaged across all simulated data sets, as a function of number of samples per trial (Figure~\ref{fig:ori_results}C,F). This analysis confirms the same pattern: fixed sampling makes large errors in estimating $\eta$ with fewer than $20$ samples, and for $\lapse$ it requires as many as $100$ samples per trial to become approximately unbiased. IBS outperforms fixed sampling for both parameters and any number of samples, and even with as few as 2 or 3 repeats comes close to matching the RMSE of exact maximum-likelihood inference. 

\subsection{Change localization}
\label{sec:change}

\begin{figure}[htp]
  \includegraphics[width=5.2in]{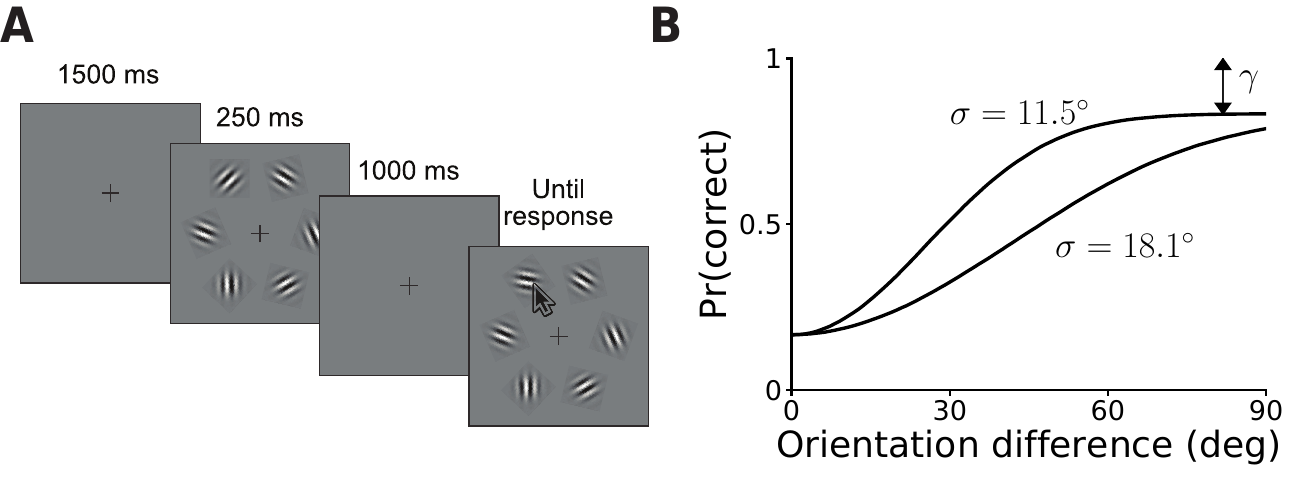}
  \vspace{-0.5em}
\caption{\textbf{A.} Trial structure of the simulated change localization task. While the participant fixates on a cross, $6$ oriented patches appear for $250$ ms, disappear and the re-appear after a delay. In the second display, one patch will have changed orientation, in this example the top left. The participant indicates with a mouse click which patch they believe changed. 
\textbf{B.} The generative model is fully characterized by the proportion correct as function of model parameters and circular distance between the orientations of the changed patch in its first and second presentation (see text). Here we plot this curve for two values of $\eta \equiv \log \sigma$. In both curves, $\lapse=0.2$. We can read off $\eta$ from the slope and $\lapse$ from the asymptote.}
\label{fig:change_task_model}
\end{figure}

The second problem we consider is a typical `change localization' task (see Figure~\ref{fig:change_task_model}A), in which participants observe a display of $6$ oriented patches, and after a short inter-stimulus interval, a second display of $6$ patches~\citep{van2012variability}. Of these patches, $5$ are identical between displays and one denoted by $c \in \{1, \ldots, 6 \}$ will have changed orientation. The participant responds by indicating which patch they believe changed orientation. Here, on each trial the stimulus $\s$ is a vector of 12 elements corresponding to a vector of orientations (in degrees) of the six patches in the first display, concatenated with the vector of orientation of the six patches in the second display. The response $r \in \{1, \dots, 6\}$ is the patch reported by the participant.

For each dataset, we simulated $N = 400$ trials. On each trial, the patches on the first display are all independently drawn from a uniform distribution $\text{Uniform[0,360]}$. For the second display, we randomly select one of the patches and change its orientation by an amount drawn from a von Mises distribution centered at $0^\circ$ with concentration parameter $\kappa_\text{s}=1$. A von Mises distribution is the equivalent of a Gaussian distribution in circular space, and the concentration parameter is monotonically related to the precision (inverse variance) of the distribution. Note that, for mathematical convenience (but without loss of generality) we assume that patch orientations are defined on the whole circle, whereas in fact they are defined on the half-circle $[0^\circ, 180^\circ)$.

The generative model assumes that participants independently measure the orientation of each patch in both displays. For each patch, the measurement distribution is a von Mises centered on the true orientation with concentration parameter $\kappa$, representing sensory precision. 
The participant then selects the patch for which the absolute circular difference of the measurements between the first and second display is largest. This model too includes a lapse rate $\lapse \in (0,1]$, the probability with which the participant guesses uniformly randomly across responses.

Since thinking in terms of concentration parameter is not particularly intuitive, we reparametrize participants' sensory noise as $\eta \equiv \log \sigma \equiv -\frac{1}{2}\log\kappa$, since in the limit $\kappa\gg 1$, the von Mises distribution with concentration parameter $\kappa$ tends to a Gaussian distribution with standard deviation $\sigma = \frac{1}{\sqrt{\kappa}}$. The model has then two parameters, $\vtheta = (\eta, \lapse)$. 

We can express the trial likelihood for the change localization model in an integral form that does not have a known analytical solution (see Appendix \ref{sec:change_details} for a derivation). We can, however, evaluate the integral numerically, which can take a few seconds for a high-precision likelihood evaluation across all trials in a dataset. The key quantity in the computation of the trial likelihood is $\Delta_s^{(c)}$, the difference in orientation between the changed stimulus at position $c$ between the first and second display. We plot the probability of a correct response, $\Pchange$, as a function of $\Delta_s^{(c)}$ in Figure~\ref{fig:change_task_model}B. As expected, the probability of a correct response increases monotonically with the amount of change, with the slope being modulated by sensory noise and the asymptote by the lapse rate (but also by the sensory noise, for large noise, as we will discuss later).
For more details about the numerical experiments, see Appendix~\ref{sec:change_details}.

\begin{figure}[htp]
  \centering
  \includegraphics[width=5.2in]{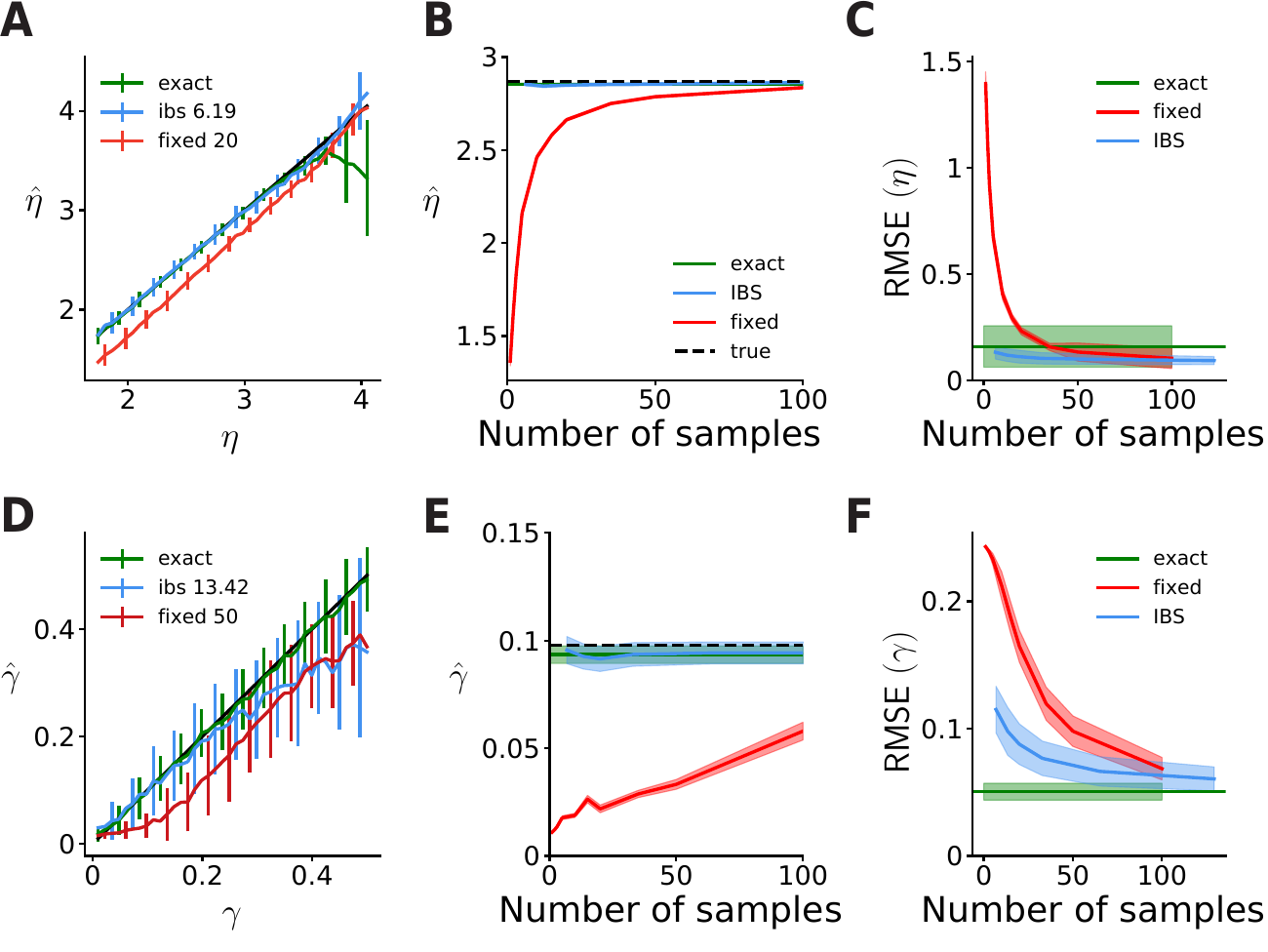}
  \vspace{-0.5em}
\caption{Same as Figure~\ref{fig:ori_results}, for the change localization experiment and estimates of $\eta \equiv \log \sigma$ and $\lapse$. In panel \textbf{A} we show results for $R = 1$ and $M = 20$; in panel \textbf{D}, $R = 2$ and $M = 50$.}
\label{fig:change_results}
\end{figure}

In Figure~\ref{fig:change_results}, we compare the performance of IBS, fixed sampling and the `exact' log-likelihood evaluated through numerical integration. As before, IBS estimates both $\eta$ and $\lapse$ more accurately with fewer samples than fixed sampling (Figure~\ref{fig:change_results}A,D). As an example, we show IBS with $R = 1$ repeats and fixed sampling with $M = 20$ or $M = 50$; the full results with all tested values of $R$ and $M$ are reported in Figure \ref{fig:vstm_complete_results} in the Appendix.

Interestingly, maximum-likelihood estimation via the `exact' method provides biased estimates of $\eta$ when the noise is high. This is because sensory noise and lapse become empirically non-identifiable for large $\eta$, as large noise produces a nearly-flat response distribution, which is indistinguishable from lapse. 
For these particular settings of $\vtheta_\text{true}$, due to the interaction between noisy log-likelihood evaluations and the optimization method, IBS and fixed sampling perform better at recovering $\eta$ than the ‘exact' method, but it does not necessarily hold true in general.
Issues of identifiability can be ameloriated by using Bayesian inference instead of maximum-likelihood estimation \citep{acerbi2014framework}.

In Figure~\ref{fig:change_results}B,E, we show the estimates of fixed sampling and IBS for simulated data with $\eta_\text{true} \equiv \log\sigma_\text{true}= \log 17.2^{\circ}$ and $\lapse_\text{true}=0.1$, and find that fixed sampling substantially underestimates $\eta$ when using less then $50$ samples, and underestimates $\lapse$ even with $100$ samples per trial. By contrast, IBS produces parameter estimates with relatively little bias and standard deviation close to that of exact maximum-likelihood estimation. Finally, in Figure~\ref{fig:change_results}C,F we show that IBS has lower RMSE than fixed sampling for both parameters when compared on equal terms of number of samples.

\subsection{Four-in-a-row game}
\label{sec:fourinarow}

\begin{figure}[htp]
  \includegraphics[width=5.2in]{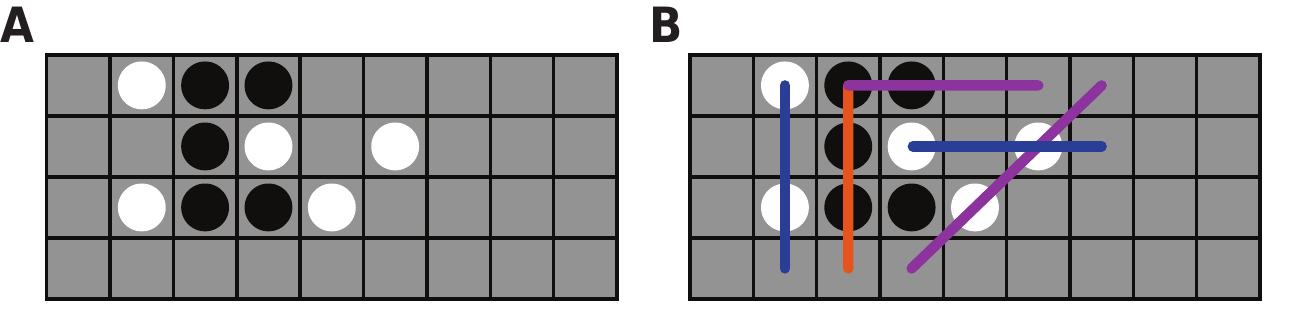}
  \vspace{-0.5em}
\caption{\textbf{A.} Example board configuration in the 4-in-a-row task, in which two players alternate placing pieces (white or black circles) on a $4$-by-$9$ board (gray grid), and the first player to get 4 pieces in a row wins. In this example, the black player can win by placing a piece on the square on the bottom row, third column. 
\textbf{B.} Illustration of features used in the value function of the heuristic search model (Equation~\ref{eq:valuefunction}). For details on the model, see Appendix~\ref{sec:fourinarow_details} and \citet{van2016people}.}
\label{fig:fourinarow_task_model}
\end{figure}

The third problem we examine is a complex sequential decision-making task, a variant of tic-tac-toe in which two players compete to place $4$ pieces in a row, column or diagonal on a $4$-by-$9$ board (see Figure~\ref{fig:fourinarow_task_model}A). In previous work, \citet{van2016people} showed that people's decision-making process in this game can be modeled accurately as \emph{heuristic search}. A heuristic search algorithm makes a move in a given board state by searching through a decision tree of move sequences starting at that board state for a number of moves into the future. To decide which candidate future moves to include in the tree, the algorithm uses a \emph{value function} defined as 
\begin{equation}\label{eq:valuefunction}
V(\text{board},\text{move}) = \sum_{i=1}^{n_\text{f}}w_if_i(\text{board},\text{move}) + \mathcal{N}(0,\sigma^2),
\end{equation}
in which $f_i$ denotes a set of $n_\text{f}$ features (i.e., configurations of pieces on the board, such as `three pieces on a row of the same color'; see Figure~\ref{fig:fourinarow_task_model}B), $w_i \in \mathbb{R}$ the corresponding feature weights, and $\sigma > 0$ is a model parameter which controls value noise. As before, we parameterize the model with $\eta \equiv \log\sigma$. 

The interpretation of this value function is that moves which lead to a high value $V(\text{board},\text{move})$ are given priority in the search algorithm, and the model is more likely to make those moves. As a heuristic to reduce the search space, any moves for which the value $V(\text{board},\text{move})$ is less than that of the highest-value move minus a \emph{threshold} parameter $\thresh > 0$ are \emph{pruned} from the tree and never considered as viable options. Finally, when evaluating $V(\text{board},\text{move})$, the model stochastically omits features from the sum $\displaystyle\sum_{i=1}^{n_\text{f}}w_if_i$; the probability for any feature to be omitted (or \emph{dropped}) is independent with probability $\delta \in [0,1]$ (the \emph{drop rate}). \citet{van2016people} considered various heuristic search models with different feature sets, and estimated the value of feature weights $w_i$ as well as the size (number of nodes) of the decision tree based on human data. Here, we consider a reduced model in which the feature identity $f_i$, feature weights $w_i$ and tree size are fixed (see Appendix~\ref{sec:fourinarow_details} for their values). Thus, the current model has three parameters, $\vtheta = (\eta,\thresh,\delta)$.

Note that even though the 4-in-a-row task is a sequential game, the heuristic search model makes an independent choice on each move, with the `stimulus' $s$ on each trial being the current board state. Hence, the model satisfies the conditional independence assumptions of Equations \ref{eq:likelihood_ind} and \ref{eq:simulator_ind}. Note also that, even though the heuristic search algorithm can be specified as a generative `simulator' which we can query to make moves in any board position, we have no way of calculating the distribution over its moves, since this would require integrating over all possible trees it could build, features which may be dropped, and realizations of the value noise. Therefore, we are in the scenario in which log-likelihood estimation is only possible by simulation, and we cannot compare the performance of fixed sampling or IBS to any `exact' method. 

To generate synthetic data sets for the 4-in-a-row task, we first compiled a set of $5482$ board positions which occurred in human-versus-human play \citep{van2016people}. For each data set, we then randomly sampled $N = 100$ board positions without replacement which we used as `stimuli' for each trial, and sampled a move from the heuristic search algorithm for each position to use as `responses'. For more details about the numerical experiments, see Appendix~\ref{sec:fourinarow_details}.

\begin{figure}[htp]
  \centering
  \includegraphics[width=5.2in]{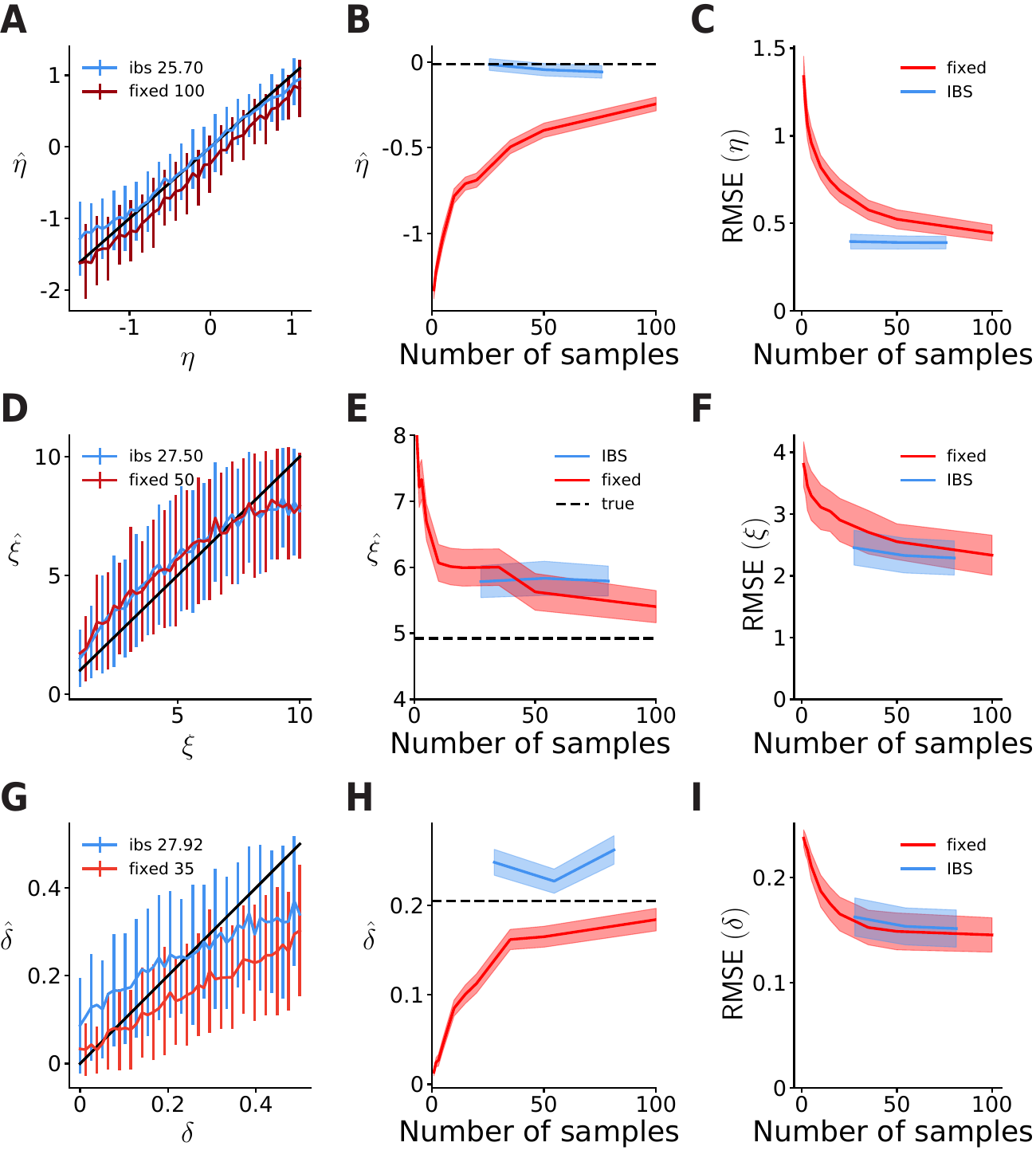}
  \vspace{-0.5em}
\caption{Same as Figures~\ref{fig:ori_results} and~\ref{fig:change_results}, for the 4-in-a-row experiment and estimates of the value noise $\eta \equiv \log\sigma$, pruning threshold $\thresh$ and feature drop rate $\delta$. In panel \textbf{A} we show results for $R = 1$ and $M = 100$; in panel \textbf{D}, $R = 1$ and $M = 50$; in panel \textbf{G}, $R = 1$ and $M = 35$.}
\label{fig:fourinarow_results}
\end{figure}

In Figure~\ref{fig:fourinarow_results}, we perform the same tests as before, comparing fixed sampling and IBS, but lacking any `exact' estimation method. Due to the high computational complexity of the model, we only consider IBS with up to $R = 3$ repeats, corresponding to $\sim 80$ samples. The full results with all tested values of $R$ and $M$ are reported in Figure \ref{fig:fourinarow_complete_results} in the Appendix. As a specific example for Figure~\ref{fig:fourinarow_results}B,E,H we show the estimates of fixed sampling and IBS for simulated data with $\eta_\text{true} \equiv \log\sigma_\text{true}= \log 1$, pruning threshold $\thresh_\text{true} = 5$ and $\delta_\text{true}=0.2$. 

Fixed sampling underestimates the value noise $\eta$, even when using $M=100$ samples, whereas IBS estimates it accurately with $4$ times fewer samples (Figure~\ref{fig:fourinarow_results}A). 
This bias of fixed sampling gets worse with fewer samples (Figure~\ref{fig:fourinarow_results}B), and overall, IBS outperforms fixed sampling when compared on equal terms (Figure~\ref{fig:fourinarow_results}C). The same holds true for the pruning threshold $\thresh$. IBS estimates $\thresh$ about equally well as fixed sampling, but with about half as many samples (Figure~\ref{fig:fourinarow_results}D), fixed sampling is severely biased when using too few samples (Figure~\ref{fig:fourinarow_results}E) and overall, IBS outperforms fixed sampling. 

The results are slightly more complicated for the feature drop rate $\delta$. As before, fixed sampling produces strongly biased estimates of $\delta$ with up to $35$ samples (Figure~\ref{fig:fourinarow_results}G), and the bias increases when using fewer samples (Figure~\ref{fig:fourinarow_results}H). However, for this parameter IBS is also biased, but towards $0.25$ (Figure~\ref{fig:fourinarow_results}G \& H), which is the midpoint of the `plausible' upper and lower bounds used as reference by the optimization algorithm (see Appendix \ref{sec:fourinarow_details} for details). This bias can be interpreted as a form of regression towards the mean; likely a by-product of the optimization algorithm struggling with a low signal-to-noise ratio for this parameter and these settings (i.e., a nearly flat likelihood landscape for the amount of estimation noise on the log-likelihood). The negative bias of fixed sampling helps to reduce its variance in the low-$\delta$ regime, and therefore in terms of RMSE, fixed sampling performs similarly to IBS for this parameter (Figure~\ref{fig:fourinarow_results}I). 

\subsection{Log-likelihood loss}
\label{sec:loglikeloss}

In the previous sections, we have analyzed the bias and error of different estimation methods when recovering the generating model parameters in various scenarios. Another important question, crucial for model selection, is how well different methods are able to recover the true maximum log-likelihood. The ability to recover the true parameters and the true maximum log-likelihood are related but distinct properties because, for example, a relatively flat likelihood landscape could yield parameter estimates very far from ground truth, but still afford recovery of a value of the log-likelihood close to the true maximum.
We recall that differences in log-likelihood much greater than one point are worrisome as they might significantly affect the outcomes of a model comparison \citep{kass1995bayes,jeffreys1998theory,anderson2002model}.

To compute the \emph{log-likelihood loss} of a method for a given data set, we estimate the difference between the `exact' log-likelihood evaluated at the `true' maximum-likelihood solution (as found after multiple optimization runs) and the `exact' log-likelihood of the solution returned by the multi-start optimization procedure for a given method, as described in Section \ref{sec:procedure}. In terms of methods, we consider IBS and fixed-sampling with different amounts of samples. We perform the analysis for the two scenarios, orientation discrimination (Section \ref{sec:ori}) and change localization (Section \ref{sec:change}), for which we have access to the exact likelihood, either analytically or numerically.

\begin{figure}[htp]
  \centering
  \includegraphics[width=5.2in]{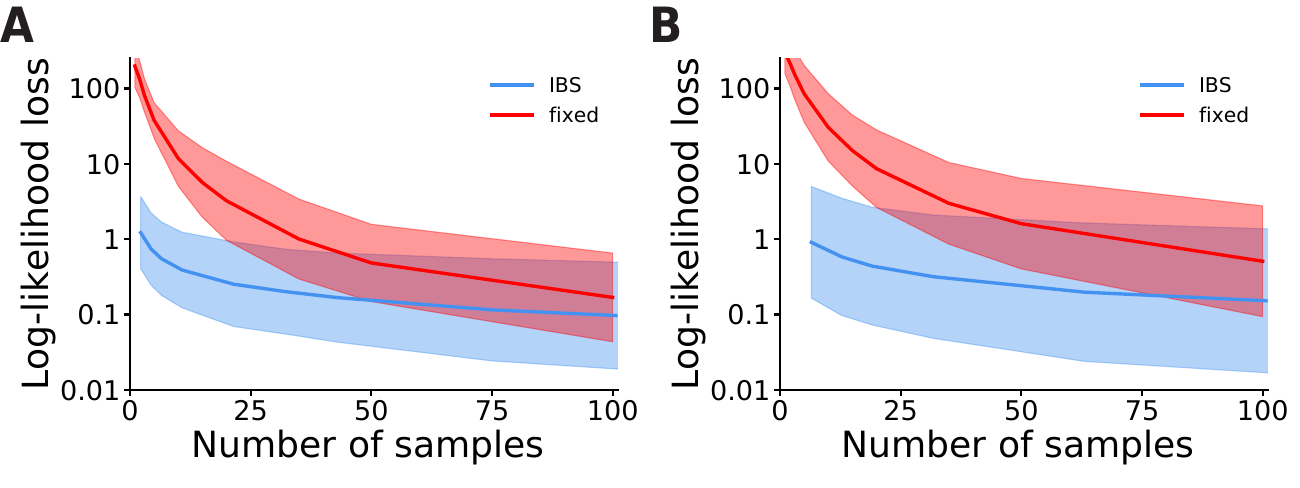}
  \vspace{-0.5em}
\caption{\textbf{A.} Log-likelihood loss with respect to ground truth, as a function of number of samples, for the orientation discrimination task. Lines are mean $\pm 1$ standard deviation (in log space) across $120$ generating parameter values, with $100$ simulated datasets each. \textbf{B.} Log-likelihood loss for the change localization task ($80$  generating parameter values).}
\label{fig:logloss}
\end{figure}

The results in Figure \ref{fig:logloss} show that IBS, even with only a few repeats, is able to return solutions which are very close to the true maximum-likelihood solution in terms of log-likelihood (within 1-2 points); whereas fixed sampling remains strongly biased (log-likelihood loss $\gg$ 1) even with large number of samples, being thus at risk of inducing wrong inferences in model selection.
Note that our analyses of the loss are based on the `exact' log-likelihood values evaluated at the solution returned by the optimization procedure. In practice, we would not have access to the `exact' log-likelihood at the solution; but its value can be estimated up to the desired precision with IBS, by taking multiple repeats at the returned solution (see Section \ref{sec:multifidelity}).

Finally, the results in Figure \ref{fig:logloss} also display clearly that, while IBS is unbiased in estimating the log-likelihood \emph{for a given parameter setting} $\vtheta$, the maximum-likelihood procedure per se will have some error. Due to estimation noise and specific features of the data, model, and stochastic optimization method at hand, the returned solution will rarely be the true maximum-likelihood solution, and thus, by definition, the value of the log-likelihood at the solution will \emph{underestimate} the true value of the maximum log-likelihood. Still, Figure \ref{fig:logloss} shows that the underestimation error, at least in the IBS case, tends to be acceptable, as opposed to the large errors obtained with fixed sampling.

\subsection{Summary}

The results in this section demonstrate that in realistic scenarios, fixed sampling with too few samples causes substantial biases in parameter and maximum log-likelihood estimates, whereas inverse binomial sampling is much more accurate and robust to the number of samples used. Across all $3$ models and all parameters, IBS yields parameter estimates with little bias and RMSE, close to that of `exact' maximum-likelihood estimation, even when using only a handful of repeats ($R$ between 1 and 5). Conversely, fixed sampling yields substantially biased parameter estimates when using too few samples per trial, especially for parameters which control decision noise, such as measurement noise and lapse rates in the two perceptual decision-making tasks, and value noise in the 4-in-a-row task. Moreover, for the two models for which we have access to `exact' log-likelihood estimates, we found that IBS is able to recover maximum-likelihood solutions close to the true maximum log-likelihood, whereas fixed sampling remains severely biased even for many samples.
 
It is true that, given a \emph{large enough} number of samples, fixed sampling is eventually able to recover most parameters and maximum log-likelihood values with reasonable accuracy. However, we have seen empirically that the number of samples required for reliable estimation varies between tasks, models and parameters of interests. For tasks and models where an exact likelihood or a numerical approximation thereof is unavailable, such as the problem we examined in Section \ref{sec:fourinarow}, this limitation renders fixed sampling hardly usable in practice. By contrast, IBS automatically chooses the number of samples to allocate to the problem.

Finally, for complex models with a large response space, accurate parameter estimation with fixed sampling will require many more samples per trial than are feasible given the computational time needed to generate them. Therefore, in such scenarios accurate and efficient parameter estimation is only possible with IBS.

\section{Discussion}
\label{sec:discussion}

In this work, we presented inverse binomial sampling (IBS), a method for estimating the log-likelihood of simulation-based models given an experimental data set. We demonstrated that estimates from IBS are uniformly unbiased, their variance is uniformly bounded, and we introduced a calibrated estimator of the variance. IBS is sample-efficient and, for the purpose of maximum-likelihood estimation, combines naturally with gradient-free optimization algorithms that handle stochastic objective functions, such as Bayesian Adaptive Direct Search (BADS; \citealp{acerbi2017practical}).
We compared IBS to fixed sampling and showed that the bias inherent in fixed sampling can cause researchers to draw false conclusions when performing model selection. Moreover, we showed in three realistic scenarios of increasing complexity that maximum-likelihood estimation of model parameters is more accurate with IBS than with fixed sampling with the same average number of samples. 

In the rest of this section, we discuss additional applications of IBS, possible extensions, and give some practical usage recommendations.

\subsection{Additional applications}
\label{sec:applications}

We developed inverse binomial sampling for log-likelihood estimation of models with intractable likelihoods, for the purpose of model comparison or fitting model parameters with maximum-likelihood estimation, but IBS has other practical uses.

\subsubsection*{Checking analytical or numerical log-likelihood calculations}

We presented IBS as a solution for when the log-likelihood is intractable to compute analytically or numerically. However, even for models where the log-likelihood could be specified, deriving it can be quite involved and time-consuming, and mistakes in the calculation or implementation of the resulting equations are not uncommon. In this scenario, IBS can be useful for:
\begin{itemize}
\item quickly prototyping (testing) of new models, as writing the generative model and fitting it to the data is usually much quicker than deriving and implementing the exact log-likelihood;
\item checking for derivation or implementation mistakes, as one can compare the \emph{supposedly} `exact' log-likelihood against estimates from IBS (on real or simulated data);
\item assessing the quality of numerical approximations used to calculate the log-likelihood, for example when using methods such as adaptive quadrature for numerical integration~\citep{press1992numerical}.
\end{itemize}

\subsubsection*{Estimating entropy and other information-theoretic quantities}
We can also use inverse binomial sampling to estimate the entropy of an arbitrary discrete probability distribution $\Pr(x)$, with $x \in \Omega$, a discrete set (see, e.g., \citealp{cover2012elements}, for an introduction to information theory). To do this, we first draw a sample $x$ from the distribution, then use IBS to estimate $\log \Pr(x)$. The first sample and the samples in IBS are independent, and therefore we can calculate the expected value of the outcome of IBS,
\begin{equation}
\mathbb{E}\left[\Libs\right]=\mathbb{E}_{x\sim \Pr(\cdot)}\left[\log \Pr(x)\right]=\sum_{x\in \Omega}\Pr(x)\log \Pr(x),
\end{equation}
which is the definition of the negative entropy of $\Pr(x)$. 

We can use this technique to estimate the entropy of the predicted response distribution of a generative model with a given parameter vector on any trial. For example, such quantity could be used in a behavioral model to test for the generalized Hick-Hyman law, that states that reaction time is proportional to the entropy of the available choices~\citep{hyman1953stimulus}. Moreover, we can generalize the method to estimate the cross-entropy between two distributions (sample from one, estimate log-likelihood with the other), or the Kullback-Leibler divergence between distributions. 
We note that all the estimates of these quantities are also uniformly unbiased.\footnote{The lack of bias in entropy estimates by IBS may be surprising in light of a theorem stating that uniformly unbiased estimators of the entropy given a finite set of samples cannot exist \citep{paninski2003estimation}. This theorem does not apply to IBS since its sample size is a stochastic variable. It does, however, prove that one cannot estimate entropy (or similar information-theoretic quantities) with fixed sampling.}

\subsection{Bayesian inference}

In this paper we focused on maximum-likelihood estimation, but another common approach to parameter estimation is Bayesian inference~\citep{gelman2013bayesian}. Bayesian inference has the goal of computing the \emph{posterior distribution} of the parameters given the observations, computed as
\begin{equation} \label{eq:bayes}
p(\vtheta | \data) = \frac{\Pr(\data | \vtheta) p(\vtheta)}{\mathcal{Z}} \qquad \text{ with } \mathcal{Z} \equiv \int d \Pr(\data | \vtheta) p(\vtheta) d\vtheta,
\end{equation}
where $\Pr(\data | \vtheta)$ is the likelihood, $p(\vtheta)$ the prior density of the parameters (typically assumed continuous), and $\mathcal{Z}$ the normalization constant, known as the \emph{evidence} or \emph{marginal likelihood}, a quantity used for Bayesian model selection due to a number of desirable properties~\citep{mackay2003information}.
Since $\mathcal{Z}$ is often hard to compute, many (approximate) Bayesian inference techniques are able to calculate the posterior distribution by having access only to the \emph{unnormalized} posterior, or joint distribution $\Pr(\data | \vtheta) p(\vtheta)$; or equivalently to the log joint $\L(\vtheta) + \log p(\vtheta)$. We see then that IBS could be used to perform Bayesian inference of likelihood-free models by providing a means to compute the log-likelihood in the log joint distribution (the prior is assumed to be a simple distribution which we can express in closed form). 

In Appendix \ref{sec:bayesian}, we describe how several approaches to approximate Bayesian inference could be used in conjunction with the unbiased log-likelihood estimates provided by IBS: Markov Chain Monte Carlo \citep{hastings1970monte,brooks2011handbook}; variational inference \citep{jordan1999introduction,ranganath2014black}; and Gaussian process surrogate methods \citep{kandasamy2015bayesian,jarvenpaa2019parallel}, including Variational Bayesian Monte Carlo (VBMC; \citealt{acerbi2018variational,acerbi2019exploration}). In particular, \citet{acerbi2020variational} demonstrates the effectiveness of IBS, combined with VBMC, for robust and sample-efficient Bayesian inference, using a variety of models from computational and cognitive neuroscience.

Finally, note that the techniques in this paper can be easily applied to maximum-a-posteriori (MAP) estimation -- which is not quite Bayesian inference, but more like a regularized form of maximum-likelihood, that still yields a point estimate instead of a full posterior distribution. MAP estimation is attained by simply adding the log-prior to the log-likelihood in the optimization objective, where the log-prior acts as a regularization term.

\subsection{Approximate IBS for continuous responses}
\label{sec:ABC}

So far, we have assumed that the space of possible responses is discrete. This assumption is necessary since, for continuous responses, the probability that a sample from the generative model exactly matches an observed response is zero (technically, \emph{near}-zero since any computer implementation of a real number is finite). For this reason, IBS will never terminate, or at least not within a physically sensible time scale.

A simple approach to make continuous responses discrete is via binning the response space. Alternatively, we recommend an approach inspired by Approximate Bayesian Computation (ABC; \citealp{beaumont2002approximate}), which we call Approximate IBS (AIBS). Given a metric $D(\cdot, \cdot)$ to measure distance in response space, and a tolerance threshold $\varepsilon > 0$, we can use IBS to estimate
\begin{equation} \label{eq:abc}
\L_{\varepsilon}\left(\vtheta\right)=\sum_{i=1}^N\log \frac{\Pr \left(D(\tilde{\r}_i,\r_i)\leq \varepsilon\lvert \s_i,\vtheta\right)}{\left|B_\varepsilon(\r_i)\right|},
\end{equation}
where the $\tilde{\r}_i$ are responses drawn from the generative model, and $|B_\varepsilon(\r_i)|$ denotes the volume of the set of responses whose distance from $\r_i$ is no more than $\varepsilon$.

The $\varepsilon$-approximate log-likelihood in \eq~\ref{eq:abc} can then be used as normal for maximum-likelihood estimation or Bayesian inference.
As $\varepsilon\rightarrow 0$, the approximate likelihood tends to the true likelihood, under some regularity conditions which we leave to explore for future work (see \citealt{prangle2017adapting} for a similar proof for ABC). However, the expected number of samples used by IBS diverges in that limit, so in practice there is a lower bound for $\varepsilon$ that is feasible and one needs to extrapolate to the $\varepsilon=0$ limit, or be satisfied to perform inference with an $\varepsilon$-approximate likelihood.

The common idea between AIBS and ABC is that they both use a distance metric to judge similarity between simulated samples and data. However, ABC commonly bases the comparison on \emph{summary statistics} of the data (which may not be \emph{sufficient} statistics, and thus not capture all aspects of the data); whereas AIBS uses the full responses. Secondly, ABC in practice requires dedicated algorithms to perform parameter estimation and inference (basic techniques, such as rejection sampling, can be extremely inefficient); whereas AIBS simply provides a (noisy) log-likelihood, which can then be used in combination with a wider class of likelihood-based inference methods, as long as they support noisy estimates (see Appendix \ref{sec:bayesian} for some examples). We leave a further analysis of AIBS, and a comparison with other likelihood-free inference approaches, as a promising direction for future work.

\subsection{Usage recommendations}

We conclude with a number of recommendations for researchers who want to fit a model to a data set, having access only to a simulator or generative model. 
\begin{itemize}
\item First, try to derive a closed-form analytic expression for the log-likelihood of the model. If this is tractable, validate that the log-likelihood is free of implementation mistakes by comparing its output against log-likelihood estimates obtained by IBS with well-chosen test trials and model parameters. 
\item If exact analytics are intractable, find an analytical or numerical approximation, for example using variational inference or Riemannian integration, and once again validate the quality of the approximation using IBS. 
\item If the model is too complex for analytical or numerical approximations, estimate the log-likelihood using inverse binomial sampling. 
\item Finally, perform inference using the analytical, numerical, or IBS-based log-likelihood function with a sample-efficient inference algorithm, such as those based on Gaussian process surrogate modeling. For maximum-likelihood (or maximum-a-posteriori) estimation, hybrid Bayesian optimization methods have proved to be quite effective \citep{acerbi2017practical}.
\end{itemize}

\subsubsection*{Avoiding infinite loops}

One issue of IBS is that it can `hang', in the sense that the implementation of the estimator can run indefinitely, without returning an answer, if the simulator is unable to match a particularly unlikely observation. This is a natural behavior of IBS that stems from its efficiency in allocating samples, as we examined in Section \ref{sec:time}. We recommend two easy solutions to avoid infinite loops:
\begin{itemize}
\item Implement a `lapse rate' $\lapse \in (0, 1)$ in the simulator model, which represents the probability of a completely random response (typically uniform across all possible responses). The lapse rate could be fixed to a small, non-zero value (e.g., $\lapse = 0.01$), or let as a free model parameter; in which case, ensure that the \emph{minimum} lapse rate is a small, non-zero value (e.g., $\lapse_\text{min} = 0.005$).
\item Introduce an early-stopping threshold, such that IBS stops sampling when the estimated log-likelihood of the entire data set goes below a threshold $\L_\text{lower}$ (see Appendix \ref{sec:threshold}).
\end{itemize}
We implemented both of these solutions in our analyses in Section \ref{sec:experiments}.


%
%

\subsection*{Acknowledgments}

This work has utilized the NYU IT High Performance Computing resources and services. We thank Aspen Yoo for help with Figures~\ref{fig:ori_task_model}~and~\ref{fig:change_task_model} and useful comments on the manuscript, and Michael Landy for helpful discussion about the derivation of the variance of the IBS estimator.
Luigi Acerbi was partially supported by the Academy of Finland Flagship programme: Finnish Center for Artificial Intelligence (FCAI).

\bibliographystyle{APA}

\clearpage

\appendix

\part*{Supplementary Material}

\makeatletter
\renewcommand{\thefigure}{S\@arabic\c@figure}
\renewcommand{\thetable}{S\@arabic\c@table}
\renewcommand{\theequation}{S\@arabic\c@equation}
\setcounter{table}{0}
\setcounter{equation}{0}
\setcounter{figure}{0}

\section{Further theoretical analyses}

\subsection{Why inverse binomial sampling works} 
\label{sec:ibsworks}

We start by showing that the inverse binomial sampling policy described in Section \ref{sec:policies}, combined with the estimator $\hat{\L}_{\text{ibs}}$ (\eq~\ref{eq:Libs}), yields a uniformly unbiased estimate of $\log p$. This derivation follows from \citet[Theorem 4.1]{degroot1959unbiased}, adapted to our special case of estimating $\log p$ instead of an arbitrary function $f(p)$:
\begin{equation} \label{eq:ibsderivation}
\begin{split}
\mathbb{E}\left[\hat{\L}_{\text{ibs}}\right]
= &-\mathbb{E}\left[\sum_{k=1}^{K-1}\frac{1}{k}\right]
= -\mathbb{E}\left[\sum_{k=1}^{\infty}\frac{1}{k}\mathbbm{1}_{k < K}\right] \\
 = & -\sum_{k=1}^{\infty}\frac{1}{k}\mathbb{E}\left[\mathbbm{1}_{k < K}\right]
 = -\sum_{k=1}^{\infty}\frac{1}{k}\Pr\left(k < K\right) \\
 = & -\sum_{k=1}^{\infty}\frac{1}{k}(1-p)^k
 = \log p.
\end{split}
\end{equation}
The first equality is the definition of $\hat{\L}_{\text{ibs}}$ (\eq~\ref{eq:Libs}), using the notational convention that $\displaystyle\sum_{k=1}^0 = 0$. In the second equality we introduce the indicator function $\mathbbm{1}_{k < K}$ which is $1$ when $k < K$ and $0$ otherwise. The third equality follows by linearity of the expectation and the fourth directly from the definition of the indicator function. The fifth and second-to last equality uses the formula for the cumulative distribution function of a geometric variable, that is $\Pr(K \le k) = 1 - (1 - p)^k$, and thus $\Pr(k < K) = (1 - p)^k$.
The final equality is the definition of the Taylor series of $\log p$ expanded around $p=1$. Note that this series converges for all $p\in(0,1]$. 

In the derivation above, we can replace $\frac{1}{k}$ by an arbitrary set of coefficients $a_k$ and show that 
\begin{equation} \label{eq:taylor}
\mathbb{E}\left[\sum_{k=1}^{K-1}a_k\right]=\sum_{k=1}^{\infty}a_k(1-p)^k,
\end{equation}
for all $p$ for which the resulting Taylor series converges. \eq~\ref{eq:taylor} immediately proves two corollaries. First, we can use the inverse binomial sampling policy to estimate any analytic function of $p$. Second, since we can rewrite any estimator $\hat{\L}(K)$ as $\displaystyle\sum_{k=1}^{K-1} a_k$, and since Taylor series are unique, $a_k=\frac{1}{k}$ is the only choice for which $\mathbb{E}\left[\hat{\L}(K)\right]$ equals $\log p$. In other words, $\hat{\L}_{\text{ibs}}$ is the only uniformly unbiased estimator of $\log p$ with the inverse sampling policy. Therefore, it trivially is the \emph{uniformly minimum-variance unbiased estimator} under this policy, since no other unbiased estimator exist. 

\subsection{Analysis of bias of fixed sampling}
\label{sec:analysisbias}

We provide here a more formal analysis of the bias of fixed sampling.
We initially consider the estimator $\hat{\L}_\text{fixed}$ defined by \eq~\ref{eq:Lfixed} in the main text, but we will see that our arguments hold generally for any estimator based on a fixed sampling policy.

We showed in Figure~\ref{fig:bias_variance_master} that in the regime of $p \rightarrow 0$, $M \rightarrow \infty$, while keeping $pM \rightarrow \lambda$, the bias of $\hat{\L}_\text{fixed}$ tends to a master curve.
This follows since, in this limit, the binomial distribution $\text{Binom}\left(\frac{\lambda}{M},M\right)$ converges to a Poisson distribution $\text{Pois}(\lambda)$ and therefore the bias converges to 
\begin{equation}\label{eq:bias_master}
\begin{split}
 \text{Bias}\left[\hat{\L}_\text{fixed} | p\right] = & \;  
 \mathbb{E}\left[\hat{\L}_{\text{fixed}}-\log\frac{\lambda}{M}\right] \\
= &\;  \exp(-\lambda)\sum_{m=0}^{\infty}\frac{\lambda^m}{m!}\log(m+1)-\log(M+1)-\log\frac{\lambda}{M} \\
 \underset{\substack{p \rightarrow 0 \\ M \rightarrow \infty \\ pM \rightarrow \lambda}}{\longrightarrow} 
 & \; \exp(-\lambda)\sum_{m=0}^{\infty}\frac{\lambda^m}{m!}\log(m+1)-\log\lambda,
\end{split}
\end{equation}
which is the master curve in Figure~\ref{fig:bias_variance_master}.
In particular, the bias is close to zero for $\lambda\gg 1$ and it diverges when $\lambda\ll 1$, or equivalently, for $M\gg\frac{1}{p}$ and $M\ll\frac{1}{p}$, respectively.

This asymptotic behavior is not a coincidence. In fact, it is mathematically guaranteed since the Fisher information of $\text{Pois}(\lambda)$ equals $\frac{1}{\lambda}$ and the reparametrization identity for the Fisher information yields $\I_f(\log\lambda)=\lambda$ \citep{lehmann2006theory}. In the limit of $p\ll\frac{1}{M}$, which corresponds to $\lambda\ll 1$, this Fisher information vanishes and the outcome of fixed sampling simply provides zero information about $\log\lambda$ or $\log p$. Therefore, any estimates of $\log p$ are not informed by the data and instead are a function of the regularization chosen in the estimator $\hat{\L}_{\text{fixed}}$ (\eq~\ref{eq:Lfixed}). Note that the argument above does not invoke the specific form of the estimator, and therefore holds for any choice of regularization. 

\begin{figure}[htp]
	\centering
	\includegraphics[width=5.2in]{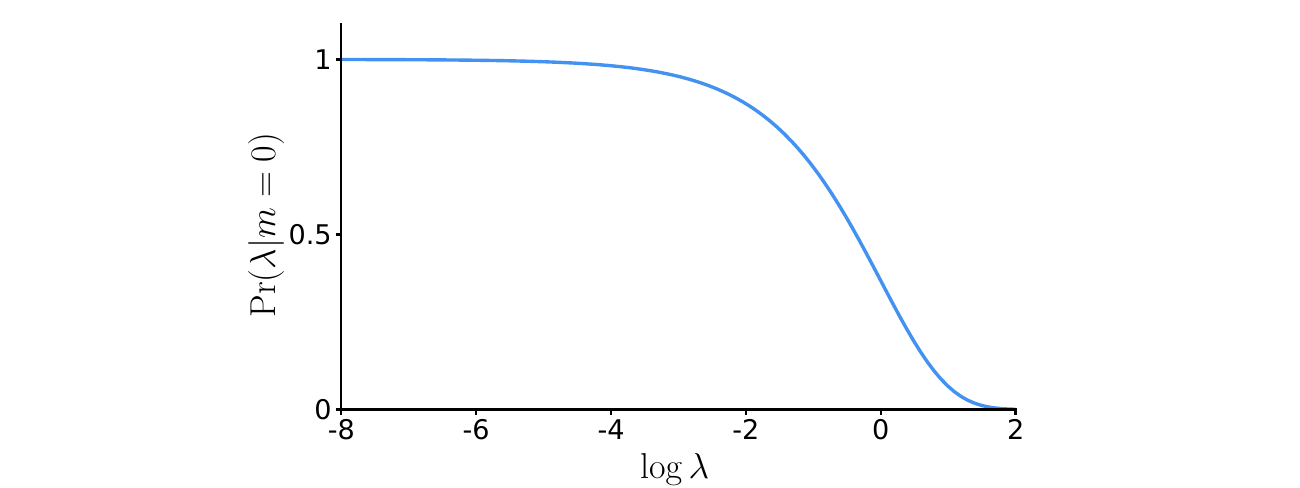}
		\vspace{-0.5em}
\caption{Likelihood function of $\log\lambda$ given that fixed sampling returns $m=0$ (none of the samples from the model match the participant's response). The likelihood is approximately flat for all $\log\lambda\leq -2$. Since $\lambda$ is defined as $\frac{p}{M}$, this implies that the posterior distribution over $p$ will be dominated by a prior rather than evidence, as quantified by the likelihood.}
\label{fig:likelihood_lambda}
\end{figure}

We can express the problem with fixed sampling more clearly using Bayesian statistics, in a formal treatment of the `gambling' analogy we presented in the main text. The `correct' belief about $\log\lambda$ given the outcome of fixed sampling ($m$) is quantified by the posterior distribution $p\left(\log\lambda\lvert m\right)$, which is a product of the likelihood $\Pr\left(m\lvert\log\lambda\right)$ and a prior $p(\log\lambda)$. In the limit $\lambda\ll 1$, the Poisson distribution converges to a Kronecker delta distribution concentrated on $m=0$. In other words, almost surely none of the samples taken by the behavioral model will match the participant's response. When $m=0$, the likelihood equals $\exp(-\lambda)$, which is mostly flat (when considered as a function of $\log\lambda$, see Figure~\ref{fig:likelihood_lambda}) for $\log\lambda\in[-\infty,-2]$ and therefore our posterior belief ought to be dominated by the prior $p(\log\lambda)$ and become independent of the data. Therefore, we once again conclude that in the limit  $p\ll\frac{1}{M}$, the fixed sampling policy provides no information to base an estimate of $\log p$ on, and it is impossible to avoid bias. 

\subsection{Derivation of IBS variance}
\label{sec:derivation_ibs_var}

In this section, we derive the expression for the variance of the IBS estimator (Equation \ref{eq:ibsvariance} in the main text).
We compute the variance of $\hat{L}_{\mathrm{ibs}}$ starting from the identity
\begin{equation*}
    \mathrm{Var}\left[\hat{L}_{\mathrm{ibs}}\right]=\mathbb{E}\left[\left(\hat{L}_{\mathrm{ibs}}\right)^2\right] - \left(\mathbb{E}\left[\hat{L}_{\mathrm{ibs}}\right]\right)^2.
\end{equation*}
We already know the second term is equal to $(\log p)^2$, but for the purpose of this derivation, and for reasons that will become clear later, we re-write it as 
\begin{equation*}
\left(\mathbb{E}\left[\hat{L}_{\mathrm{ibs}}\right]\right)^2=\left(\sum_{m=1}^{\infty}\frac{1}{m}(1-p)^m\right)^2=\sum_{m,n=1}^{\infty}\frac{1}{mn}(1-p)^{m+n}.
\end{equation*}
In order to write this equation as a power series in $1-p$, we collect terms with the same exponent together. Specifically, we re-index this double summation as a summation over all values of $n$ and $m+n$ (which we label $k$), and substitute $k-n$ for $m$. 
\begin{equation*}
\sum_{m,n=1}^{\infty}\frac{1}{mn}(1-p)^{m+n}=\sum_{k=1}^{\infty}\sum_{n=1}^{k-1}\frac{1}{(k-n)n}(1-p)^k.
\end{equation*}
Note that in the second summation over $n$ we only have to sum to $n=k-1$ since $m \equiv n-k$ has to be positive. We can carry out the internal summation over $n$ explicitly,
\begin{equation*}
\sum_{n=1}^{k-1}\frac{1}{(k-n)n}=\sum_{n=1}^{k-1}\left[\frac{1}{k(k-n)}+\frac{1}{kn}\right]=\frac{2}{k}\sum_{n=1}^{k-1}\frac{1}{n}=\frac{2}{k}H_{k-1}.
\end{equation*}
The first equality is an algebraic manipulation, the second follows by symmetry and the final equality defines $H_{k-1}$ as the $(k-1)$-th harmonic number. Therefore, we find that 
\begin{equation*}
\left(\mathbb{E}\left[\hat{L}_{\mathrm{ibs}}\right]\right)^2=2\sum_{k=1}^{\infty}\frac{H_{k-1}}{k}(1-p)^k.
\end{equation*}
To calculate $\mathbb{E}\left[\left(\hat{L}_{\mathrm{ibs}}\right)^2\right]$, we use a similar rationale as Equation \ref{eq:ibsderivation},
\begin{equation*}
\begin{split}
    \mathbb{E}\left[\left(\hat{L}_{\mathrm{ibs}}\right)^2\right]
    =& \; \mathbb{E}\left[\left(-\sum_{m=1}^{K-1}\frac{1}{m}\right)^2\right]
    = \mathbb{E}\left[\sum_{m,n=1}^{K-1}\frac{1}{mn}\right] \\
    =& \; \mathbb{E}\left[\sum_{m,n=1}^{\infty}\frac{1}{mn}\mathbbm{1}_{m < K}\mathbbm{1}_{n<K}\right] 
    = \sum_{m,n=1}^{\infty}\frac{1}{mn}\mathbb{E}\left[\mathbbm{1}_{m < K,n<K}\right]\\
    =& \sum_{m,n=1}^{\infty}\frac{1}{mn}\Pr\left(\mathbbm{1}_{m < K,n<K}\right)
    =\sum_{m,n=1}^{\infty}\frac{1}{mn}(1-p)^{\mathrm{max}(m,n)}.
\end{split}
\end{equation*}
In these equations, $\mathbbm{1}$ again denotes an indicator function, and we use the fact that the product of indicator functions for two different events is the indicator function for the joint event. Additionally, we use that the event $m < K,n<K$ is logically equivalent to $\mathrm{max}(m,n)<K$. To write this double summation as a power series, we split it into three parts: one where $m<n$, one where $m=n$ and one where $m>n$. By symmetry, the first and last part are equal, and we can write
\begin{equation*}
    \mathbb{E}\left[\left(\hat{L}_{\mathrm{ibs}}\right)^2\right] = \sum_{m=1}^{\infty}\frac{1}{mm}\left[(1-p)^{\mathrm{max}(m,m)}\right] + 2 \sum_{m=1}^{\infty}\sum_{n=1}^{m-1}\frac{1}{mn}\left[(1-p)^{\mathrm{max}(m,n)}\right].
\end{equation*}
By re-arranging some terms, and using the fact that $\max(m,m)=m$ and $\max(m,n)=m$ for all $n<m$, we can reduce this to
\begin{equation*}
    \mathbb{E}\left[\left(\hat{L}_{\mathrm{ibs}}\right)^2\right] = \sum_{m=1}^{\infty}\frac{1}{m^2}(1-p)^m + 2 \sum_{m=1}^{\infty}\left[\sum_{n=1}^{m-1}\frac{1}{n}\right]\frac{1}{m}(1-p)^m.
\end{equation*}
We can now explicitly perform the summation over $n$ in the second term and write
\begin{equation*}
    \mathbb{E}\left[\left(\hat{L}_{\mathrm{ibs}}\right)^2\right] = \sum_{m=1}^{\infty}\frac{1}{m^2}(1-p)^m + 2 \sum_{m=1}^{\infty}\frac{H_{m-1}}{m}(1-p)^m.
\end{equation*}
Finally, putting everything together, we obtain
\begin{eqnarray*}
\mathrm{Var}\left[\hat{L}_{\mathrm{ibs}}\right]&=&\mathbb{E}\left[\left(\hat{L}_{\mathrm{ibs}}\right)^2\right] - \left(\mathbb{E}\left[\hat{L}_{\mathrm{ibs}}\right]\right)^2 \\&=& \sum_{m=1}^{\infty}\frac{1}{m^2}(1-p)^m + 2 \sum_{m=1}^{\infty}\frac{H_{m-1}}{m}(1-p)^m - 2\sum_{k=1}^{\infty}\frac{H_{k-1}}{k}(1-p)^k\\ &=&
\sum_{m=1}^{\infty}\frac{1}{m^2}(1-p)^m.
\end{eqnarray*}

\subsection{Estimator variance and information inequality}
\label{sec:infoinequality}

\begin{figure}[htp]
	\centering
	\includegraphics[width=5.2in]{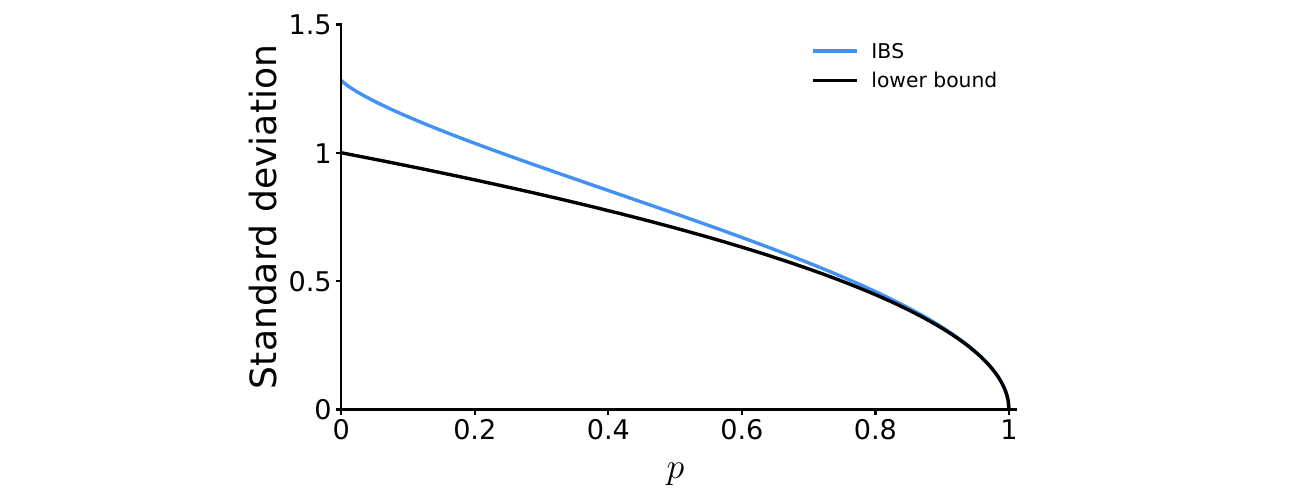}
	\vspace{-0.5em}
\caption{Standard deviation of IBS (Blue curve) and the lower bound given by the information inequality (black, see \eq~\ref{eq:information_inequality}). The standard deviation of IBS is within $30\%$ of the lower bound across the entire range of $p$. }
\label{fig:information_inequality}
\end{figure}

We proved in Section \ref{sec:ibsworks} that $\Libs$ is the minimum-variance unbiased estimator of $\log p$ given the inverse binomial sampling policy. Here we show that the estimator also comes close to saturating the \emph{information inequality}, the analogue of a Cramer-R\'ao bound for an arbitrary function $f(p)$ and a non-fixed sampling policy \citep{degroot1959unbiased},
\begin{equation}\label{eq:information_inequality}
\text{Std}(\hat{f}\lvert p)\geq\sqrt{\frac{p(1-p)}{\mathbb{E}\left[K\lvert p\right])}}\left|\frac{\mathrm{d}f(p)}{\mathrm{d}p}\right|.
\end{equation}
In our case, where $f(p)=\log p$, the information inequality reduces to $\text{Std}(\Libs)\geq \sqrt{1-p}$. In Figure~\ref{fig:information_inequality}, we plot the standard deviation of IBS compared to this lower bound. 

It may be disappointing that IBS does not match the information inequality. \citet{kolmogorov1950unbiased} showed that the only functions $f(p)$ for which the fixed sampling policy with $M$ samples allows an unbiased estimator are polynomials of degree up to $M$, and those estimators can saturate the information equality. \citet{dawson1953unbiased} and later \citet{degroot1959unbiased} showed that if an unbiased estimator of a non-polynomial function $f(p)$ exists and it matches the information inequality, it must use the inverse binomial sampling policy. Moreover, de Groot derived necessary and sufficient conditions for $f(p)$ to allow such estimators \citep{degroot1959unbiased}. Applying this argument to $f(p) = \log(p)$, the standard deviation in IBS is close (within $30\%$) to its theoretical minimum. 

\begin{figure}[htp]
	\centering
	\includegraphics[width=5.2in]{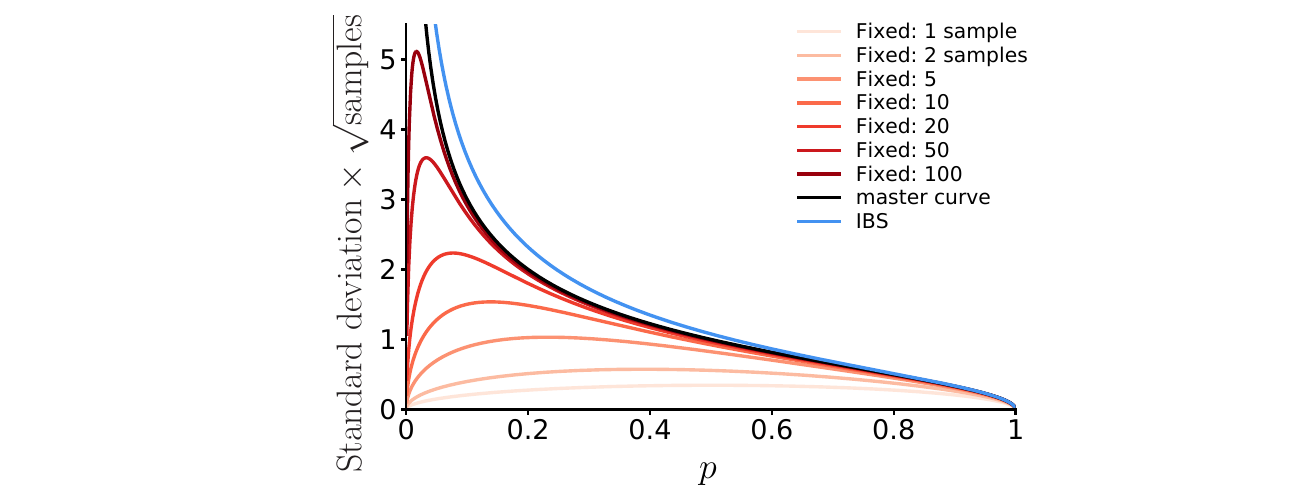}
  \vspace{-0.5em}
\caption{Standard deviation times square root of the expected number of samples drawn by IBS (blue) and fixed sampling (red), and the master curve (black) that fixed sampling converges to when $M\rightarrow\infty$.}
\label{fig:std_scaled}
\end{figure}

To compare the variance of IBS and fixed sampling on equal terms, we use the scaling behavior of $\hat{\L}_{\text{fixed}}$ as $M\rightarrow\infty$. Specifically, for fixed sampling, we plot $\sqrt{M}\times\text{Std}(\hat{\L}_{\text{fixed}})$ and for IBS we plot $\frac{1}{\sqrt{p}}\times\text{Std}(\Libs)$ (see Figure~\ref{fig:std_scaled}). With this scaling, the curves for fixed sampling again collapse onto a master curve\footnote{These curves converge pointwise on $(0,1]$ and uniformly on any interval $(\varepsilon,1]$, but not uniformly on $(0,1]$. The limits $M\rightarrow\infty$ and $p\rightarrow 0$ are not exchangeable.}. Note that repeated-sampling IBS estimators $\Libsrep{R}$ (see Section \ref{sec:multifidelity}), obtained by averaging multiple IBS estimates, overlap with the curve for regular IBS for any $R$. 

All these curves increase and diverge as $p\rightarrow 0$, reflecting the fact that estimating log-likelihoods for small $p$ is hard. The standard deviation of fixed sampling is always lower than that of IBS, especially when $p\rightarrow 0$ (specifically when $p\ll\frac{1}{M}$). In other words, fixed sampling produces low-variance estimators exactly in the range in which its estimates are biased, as guaranteed by the Cramer-R\'ao bound. However, in the large-$M$ limit, fixed sampling does saturate the information inequality, so its master curve lies below IBS. In other words, if one is able to draw so many samples that bias is no issue, then fixed sampling provides a slightly better trade-off between variance and computational time. However, in Section~\ref{sec:repeated_sampling}, we discuss an improvement to IBS which decreases its variance by a factor $2$-$20$, in which case IBS is clearly superior. 
Finally, a quantity of interest for the researcher may not be the variance of the estimator per se, but a measure of the error such as the RMSE, for which IBS is also consistently superior (see Section \ref{sec:estimator_rmse}).

\subsection{A Bayesian derivation of the IBS estimator}
\label{sec:ibs_bayesian}

In Sections \ref{sec:fixed_fail} and \ref{sec:analysisbias} we hinted at a Bayesian interpretation of the problem of estimating $\log p$. We show here that indeed we can see the IBS estimator as a Bayesian point estimate of $\log p$ with a specific choice of prior for $p$. For the rest of this section, we use $q$ to denote the likelihood of a trial (instead of $p$); that is $q$ is the parameter of the Bernoulli distribution and $\log q$ the quantity we are seeking to estimate. We changed notation to avoid confusion with expressions such as the prior probability of $q$, which is $p(q)$.

Let $K$ be the number of samples until a `hit', as per the IBS sampling policy. Following Bayes' rule, we can write the posterior over $q$ given $K$ as
\begin{equation} \label{eq:postq}
\begin{split}
p(q | K) = & \; \frac{\Pr(K | q) p(q)}{\Pr(K)} \\
 = & \; \frac{(1-q)^{K-1} q \, \text{Beta}(q; \alpha,\beta)}{\int_0^1 (1-q)^{K-1} q \, \text{Beta}(q; \alpha,\beta) dq} \\
 = & \; \frac{\Gamma(K + \alpha + \beta)}{\Gamma(\alpha+1) \Gamma(K+\beta-1)} (1 - q)^{K + \beta -2} q^{\alpha},
\end{split}
\end{equation}
where we used the fact that $\Pr(K|q)$ follows a geometric distribution, and we assumed a $\text{Beta}(\alpha,\beta)$ prior over $q$. 

In particular, let us compute the posterior mean of $\log q$ under the Haldane prior, $\text{Beta}(0,0)$ \citep{haldane1932note}. Thanks to the `law of the unconscious statistician', we can compute the posterior mean of $p(\log q | K)$ directly from \eq~\ref{eq:postq},
\begin{equation}
\begin{split}
\mathbb{E}_{p(\log q | K)}\left[\log q \right] = & \; (K-1) \int_0^1 (\log q)  (1 - q)^{K -2} dq \\
= & \; \int_0^1 (\log q) \text{Beta}(q; 1,K-1) dq \\
= & \; \psi(1) - \psi(K) \\
= & \; -\sum_{k = 1}^{K-1} \frac{1}{k},
\end{split}
\end{equation}
where the first row follows from setting $\alpha = 0$ and $\beta = 0$; it can be shown that the third row is the expectation of $\log q$ for a Beta distribution, with $\psi(z)$ the \emph{digamma function} \citep{abramowitz1948handbook}; and the last equality follows from the relationship between the digamma function and harmonic numbers, that is $\psi(n) = - \gamma + \displaystyle\sum_{k=1}^{n-1} \frac{1}{k}$, where $\gamma$ is Euler-Mascheroni constant. We also used the notational convention that $\displaystyle\sum_{k=1}^0 a_k = 0$ for any $a_k$. Note that the last row is equal to  the IBS estimator, $\Libs(K)$, as defined in \eq~\ref{eq:Libs} in the main text.

Crucially, \eq~\ref{eq:postq} shows that we can recover the IBS estimator as the \emph{posterior mean} of $\log q$ given $K$, under the Haldane prior for $q$. This interpretation allows us to also define naturally the \emph{variance} of our estimate for a given $K$, as the variance of the posterior over $\log q$,
\begin{equation}
\begin{split}
\text{Var}_{p(\log q | K)}\left[ \log q \right] = & \; \psi_1(1) - \psi_1(K),
\end{split}
\end{equation}
where $\psi_1(z)$ is the \emph{trigamma function}, the derivative of the digamma function; the equality follows from a known expression for the variance of $\log q$ under a Beta distribution for $q$.

\subsection{Estimator RMSE}
\label{sec:estimator_rmse}

In the main text and in previous comparisons we have discussed the \emph{bias} and the \emph{variance} of estimators of the log-likelihood, which are important statistical properties, but one might wonder how bias and variance combine to yield an error metric of practical relevance such as the \emph{root mean squared error} (RMSE). Crucially, this analysis depends on the number of trials $N$ (because bias and standard deviation scale differently with $N$) and on the distribution of values of the likelihood for different trials, $p_i$. 

For illustrative purposes, we took as an example the psychometric model described in Section \ref{sec:ori}, and calculated the distribution of $p_i$ for typical datasets and parameters settings. We then calculated the RMSE in estimating the total log-likelihood of a number of randomly generated datasets (sampled from the empirical distribution of $p_i$) with different number of trials; for different numbers of samples used by the IBS and fixed-sampling estimators. 

\begin{figure}[htp]
  \centering
  \includegraphics[width=5.2in]{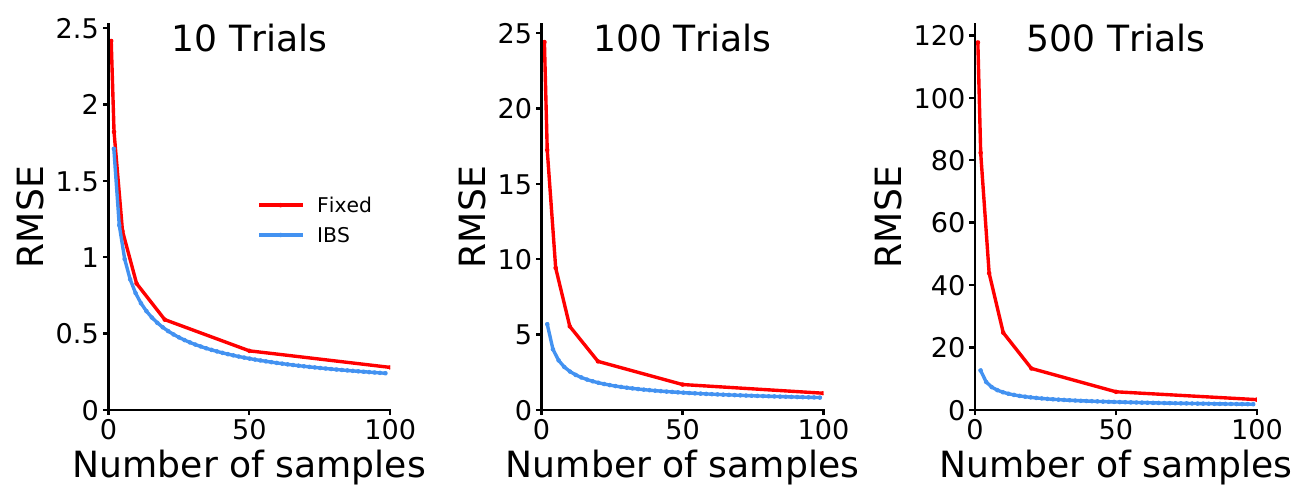}
  \vspace{-0.5em}
\caption{RMSE of the log-likelihood estimate as a function of number of samples, for the IBS and fixed-sampling estimators. Different panels display the RMSE curves for different number of trials.} 
\label{fig:psycho_rmse_example}
\end{figure}

Figure \ref{fig:psycho_rmse_example} shows that starting from even a handful of trials ($N = 10$), IBS is consistently better than fixed sampling at estimating the true value of the log-likelihood of a given parameter vector, and overwhelmingly so for a moderate number of trials ($N \ge 100$).


\section{Experimental details}

In this section, we report details for the three numerical experiments described in the main text and supplementary results.

\subsection{Orientation discrimination}
\label{sec:ori_details}

The parameters of the orientation discrimination model are the (inverse) slope, or sensory noise, represented as $\eta \equiv \log \sigma$, the bias $\mu$, and the lapse rate $\lapse$.
The logarithmic representation for $\sigma$ is a natural choice for scale parameters \cite{}(and more in general, for positive parameters that can span several orders of magnitude).

We define the lower bound (LB), upper bound (UB), plausible lower bound (PLB), and plausible upper bound (PUB) of the parameters as per Table \ref{tab:ori_params}. 
The upper and lower bounds are hard constraints, whereas the plausible bounds provide information to the algorithm to where the global optimum is likely to be, and are used by BADS, for example, to draw a set of initial points to start building the surrogate Gaussian process, and to set priors over the Gaussian process hyperparameters \citep{acerbi2017practical}. Here we also use the plausible bounds to select ranges for the parameters used to generate simulated datasets, and to initialize the optimization, as described below.

\begin{table}[ht]
\caption{Parameters and bounds of the orientation discrimination model.}
\label{tab:ori_params}
\vspace{.5em}
\begin{center}
\begin{tabular}{|cl|cc|cc|}\hline
Parameter & Description & LB    & UB  & PLB  & PUB \\
\hline 
 $\eta \equiv \log \sigma$  & Slope & $\log 0.1$ & $\log 10$ & $\log 0.1$ & $\log 5$ \\
 $\mu$ & Bias ($^\circ$)   & $-2$  & $2$ & $-1$ & $1$ \\
 $\lapse$ & Lapse  & $0.01$ & $1$  & $0.01$ & $0.2$ \\\hline
\end{tabular}
\end{center}
\end{table}

To generate synthetic data sets, we select $120$ `true' parameter settings for the orientation discrimination task as follows. We set the baseline parameter $\vtheta_0$ as $\eta = \log 2^\circ$, $\mu = 0.1^\circ$, and $\lapse = 0.1$. Then, for each parameter $\theta_j \in \{\eta, \mu, \lapse \}$, we linearly vary the value of $\theta_j$ in 40 increments from PLB$_j$ to PUB$_j$ as defined in Table \ref{tab:ori_params} (e.g., from $-1^\circ$ to $1^\circ$ for $\mu$), while keeping the other two parameters fixed to their baseline value. For each one of the 120 parameter settings $\vtheta_\text{true}$ defined in this way, we randomly generated stimuli and responses for 100 datasets from the generative model, resulting in 12000 distinct data sets for which we know the true generating parameters.

We evaluated the log-likelihood with the following methods: fixed sampling with $M$ samples, with $M \in \{1,2,3,5,10,15,20,35,50,100 \}$; IBS with $R$ repeats, with $R \in \{1,2,3,5,10,15,20,35,50\}$; and exact.
To avoid wasting computations on particularly `bad' parameter settings, for IBS we used the `early stopping threshold' technique described in Section~\ref{sec:threshold}, setting a lower bound on the log-likelihood of IBS equal to the log-likelihood of a chance model, that is $\L_\text{lower} = -N \log 2$. While this might seemingly provide an advantage to IBS with respect to Fixed sampling, note that it is simply a way to ameliorate a weakness of IBS (spending too much time on `bad' parameters vectors, which are largely inconsequential for optimization), which Fixed does not suffer from. Even so, the stopping threshold was rarely reached (2\% of evaluations).

For each data set and method, we optimized the log-likelihood by running BADS $8$ times with different starting points. We selected starting points as the points that lie on one-third or two-third of the distance between the plausible upper and lower bound for each parameter, that is all combinations of $\eta \in \{-0.998, 0.305\}$, $\mu \in \{-0.333^\circ, 0.333^\circ \}$, $\lapse \in \{ 0.073, 0.137 \}$. Each of these optimization runs returns a candidate for $\widehat{\vtheta}_\text{MLE}$.
For methods that return a noisy estimate of the log-likelihood, we then re-evaluate $\hat{\L}(\vtheta)$ for each of these $8$ candidates with higher precision (for fixed sampling, we use $10 M$ samples; for IBS, we use $10 R$ repeats). Finally, we select the candidate with highest (estimated) log-likelihood. 

When estimating parameters using IBS or fixed sampling, we enabled the `uncertainty handling' option in BADS, informing it to incorporate measurement noise into its model of the objective function. 
Note that during the optimization, the algorithm iteratively infers a single common value for the observation noise $\sigma_\text{obs}$ associated with the function values in a neighborhood of the current point \citep{acerbi2017practical}. A future extension of BADS may allow the user to explicitly provide the noise associated with each data point, which is easily computed for the IBS estimates (\eq~\ref{eq:ibsvarest} in the main text), affording the construction of a better surrogate model of the log-likelihood.

\subsubsection*{Alternative fixed sampling estimator}

In the main text, we considered the fixed-sampling estimator defined by \eq~\ref{eq:Lfixed}. We performed an additional analysis to empirically validate that our results do not depend on the specific choice of estimator for fixed sampling (as expected given the theoretical arguments in Section \ref{sec:fixed_fail}).

\begin{figure}[htp]
  \centering
  \includegraphics[width=5.2in]{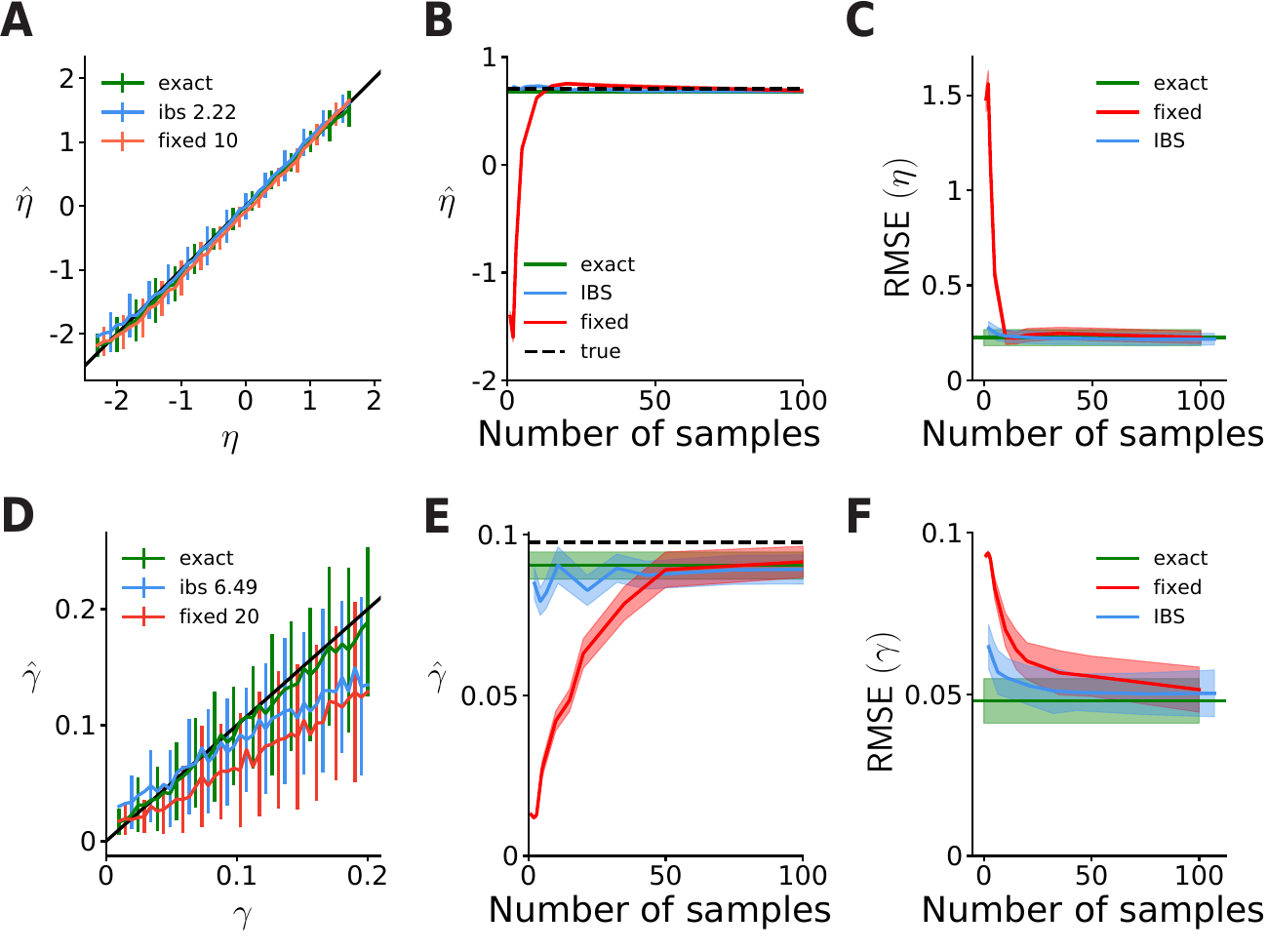}
  \vspace{-0.5em}
\caption{Same as Figure~\ref{fig:ori_results} in the main text, but for the alternative fixed-sampling estimator defined by \eq~\ref{eq:Lfixed_alt}. The results are qualitatively identical.} 
\label{fig:ori_results_b}
\end{figure}

An alternative way of avoiding the divergence of fixed sampling is to correct samples that happen to be all zeros, for example with
\begin{equation}\label{eq:Lfixed_alt}
\hat{\L}_{\text{fixed-bound}}(\x)=\log\left(   \frac{\max \left\{ m(\x), m_\text{min} \right\}}{M}\right),
\end{equation}
for some $0 < m_\text{min} < 1$, which sets a lower bound for the log-likelihood equal to $\log \left( m_\text{min} / M\right)$. 
We then performed our analyses of the orientation discrimination task using the $\hat{\L}_{\text{fixed-bound}}$ estimator with $m_\text{min} = \frac{1}{2}$. As shown in Figure \ref{fig:ori_results_b}, the results are remarkably similar to what we found using the fixed-sampling estimator $\hat{\L}_\text{fixed}$ defined by \eq~\ref{eq:Lfixed}. Finally, we also tried  $\hat{\L}_{\text{fixed-bound}}$ with a small value  $m_\text{min} = 10^{-3}$, which yielded even worse results (data not shown).

\subsubsection*{Complete parameter recovery results}

\begin{figure}[htp]
  \centering
  \includegraphics[width=5.2in]{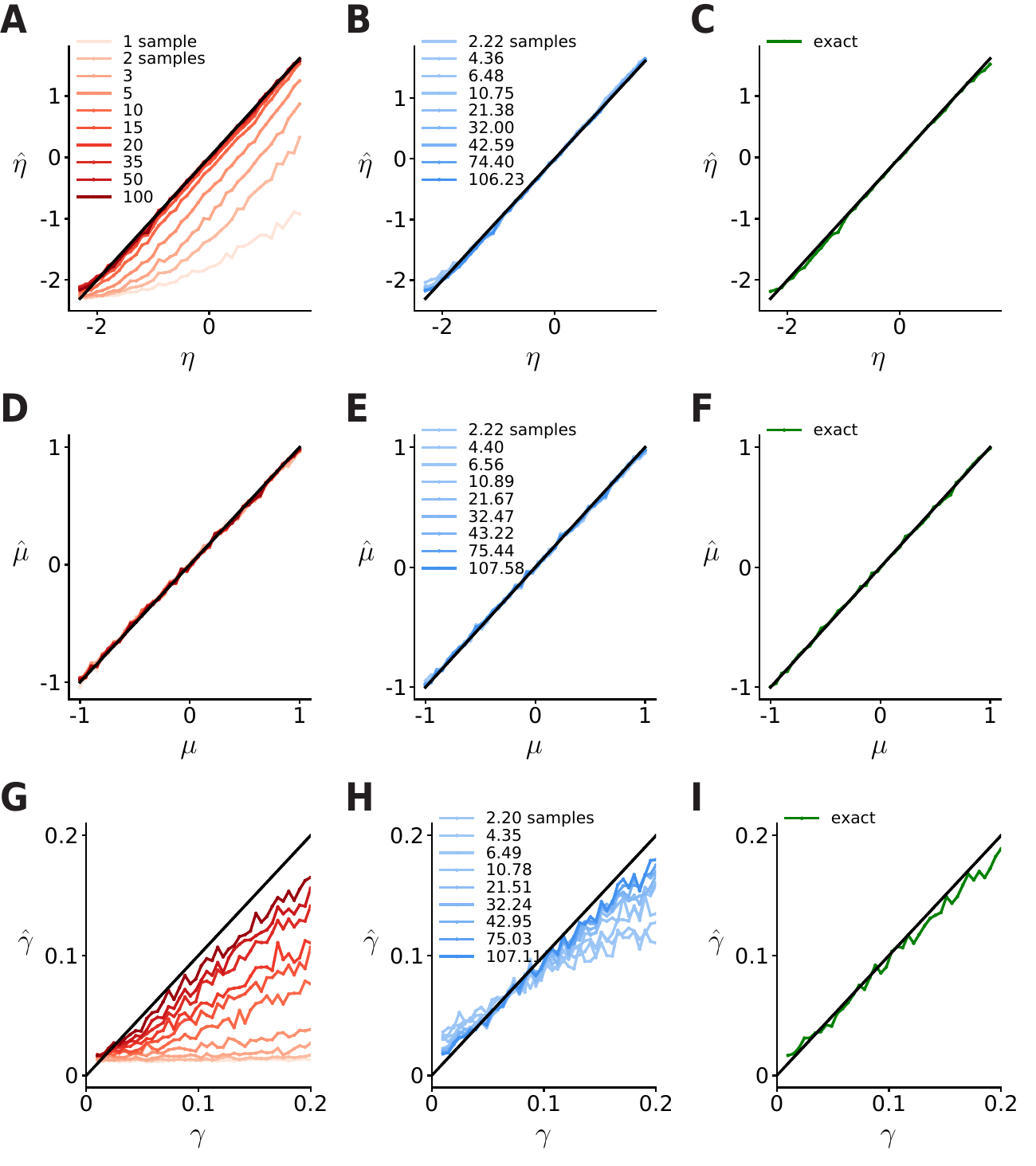}
  \vspace{-1em}
  \caption{Full parameter recovery results for the orientation discrimination model. \textbf{A.} Mean estimates recovered by fixed sampling with different number of samples. Error bars are omitted to avoid visual clutter. \textbf{B.} Mean estimates recovered by IBS with different numbers of repeats. The legend reports the average number of samples per trial that IBS uses to obtain these estimates. \textbf{C.} Mean estimate recovered using the `exact' log-likelihood function (\eq~\ref{eq:ori_equation}). \textbf{D-F} Same, for the bias parameter $\mu$. \textbf{G-I} Same, for the lapse rate $\lapse$. Overall, fixed sampling produces highly biased estimates of $\eta$ and $\lapse$, while IBS is much more accurate. 
  The bias parameter $\mu$ can be accurately estimated by either method regardless of the number of samples or repeats.}
\label{fig:psycho_complete_results}
\end{figure}

For completeness, we report in Figure~\ref{fig:psycho_complete_results} the parameter recovery results for fixed sampling, inverse binomial sampling and `exact' analytical methods for the orientation discrimination task, for all tested number of samples $M$ and IBS repeats $R$. All estimates were obtained via maximum-likelihood estimation using the Bayesian Adaptive Direct Search \citep{acerbi2017practical}, as described previously in this section.

\subsection{Change localization}
\label{sec:change_details}

First, we derive the trial likelihood of the change localization model. Assuming that the change happens at location $c \in \{1, \ldots, 6\}$, by symmetry we can write
\begin{equation}
\Pr(\text{respond }i\lvert c\text{ changed})=\begin{cases} \Pchange &\mbox{if } i=c\\ \frac{1}{5}\left(1-\Pchange\right)&\mbox{otherwise} \end{cases}
\end{equation}
where $\Delta_s^{(c)}=\left|d_\text{circ}(s_c^{(1)},s_c^{(2)})\right|$ is the absolute circular distance between the true orientations of patch $c$ in the first and second display. We can derive an expression for $\Pchange$ by marginalizing over the circular distance between the respective measurements,
\begin{equation}
\Pchange=\frac{\lapse}{6}+(1-\lapse)\int_0^{2\pi}\Pr\left(\Delta_x^{(c)}\lvert\Delta_s^{(c)}\right)\Pr\left(\forall i\neq c: \Delta_x^{(i)}\leq \Delta_x^{(c)}\lvert\Delta_s^{(i)}=0\right)\mathrm{d}\Delta_x^{(c)},
\end{equation}
where we have defined $\Delta_x^{(i)}=\left|d_\text{circ}(x_i^{(1)},x_i^{(2)})\right|$ and we suppressed the dependence on $\kappa$ to simplify the notation. The first term in this equation is the probability density function (pdf) of the circular distance between two von Mises random variables whose centers are $\Delta_s^{(j)}$ apart. The second term simplifies, since $\Delta_x^{(i)}$ for all $i\neq j$ are all independent and identically distributed. Therefore, we can rewrite this equation as 
\begin{equation}\label{eq:change_integral}
\Pchange=\frac{\lapse}{6}+(1-\lapse)\int_0^{2\pi} \Pr\left(\Delta_x^{(c)}\lvert\Delta_s^{(c)}\right)\Pr\left(\Delta\leq \Delta_x^{(c)}\right)^5\mathrm{d}\Delta_x^{(c)},
\end{equation}
where $\Delta$ is an auxiliary variable generated by taking the absolute circular difference between two von Mises random variables that are centered at $0$ with concentration parameter $\kappa$. The second term of the integrand, therefore, is the fifth power of the cumulative distribution function (cdf) of $\Delta$. We can compute the distribution of the circular distance between two von Mises random variables analytically, but the cdf is non-analytic. Moreover, the integral in \eq~\ref{eq:change_integral} is analytically intractable as well. We can, however, evaluate it numerically via trapezoidal integration (see Figure~\ref{fig:change_task_model}B). 

We now describe the settings used for maximum-likelihood estimation.
The model parameters are the sensory noise, represented as $\eta \equiv \log \sigma$ (with $\sigma = \frac{1}{\sqrt{\kappa}}$), and the lapse rate $\lapse$, with bounds defined in Table \ref{tab:change_params}.
We use the same procedure and settings for BADS as for the orientation discrimination task (see Section \ref{sec:ori_details}). For IBS, we use an early-stopping threshold of $\L_\text{lower} = -N\log 6$, and we use repeats $R \in \{1,2,3,5,10,15,20\}$ (since due to the larger response space IBS uses more samples per run). We run BADS $4$ times, with starting values of $\eta \in\{-1.535,-0.767\}$ and $\lapse\in\{0.173,0.337\}$. For data generation, we select $40$ parameter vectors with $\eta = \log 0.3$ and $\lapse$ linearly spaced from $0.01$ to $0.5$ and $40$ data sets with $\lapse=0.03$ and $\eta$ between $\log 0.1$ and $\log 1$. Again, we generate $100$ data sets for each such parameter combination.

\begin{table}[ht]
\caption{Parameters and bounds of the change localization model.}
\label{tab:change_params}
 \vspace{.5em}
\begin{center}
\begin{tabular}{|cl|cc|cc|}\hline
Parameter & Description & LB    & UB  & PLB  & PUB \\
\hline 
$\eta\equiv\log \sigma$ & Sensory noise  & $\log 0.05$ & $\log 2$ & $\log 0.1$ & $\log 1$ \\
 $\lapse$ & Lapse  & $0.01$ & $1$  & $0.01$ & $0.5$ \\\hline
\end{tabular}
\end{center}
\end{table}

\subsubsection*{Complete parameter recovery results}

\begin{figure}[htp]
  \centering
  \includegraphics[width=5.2in]{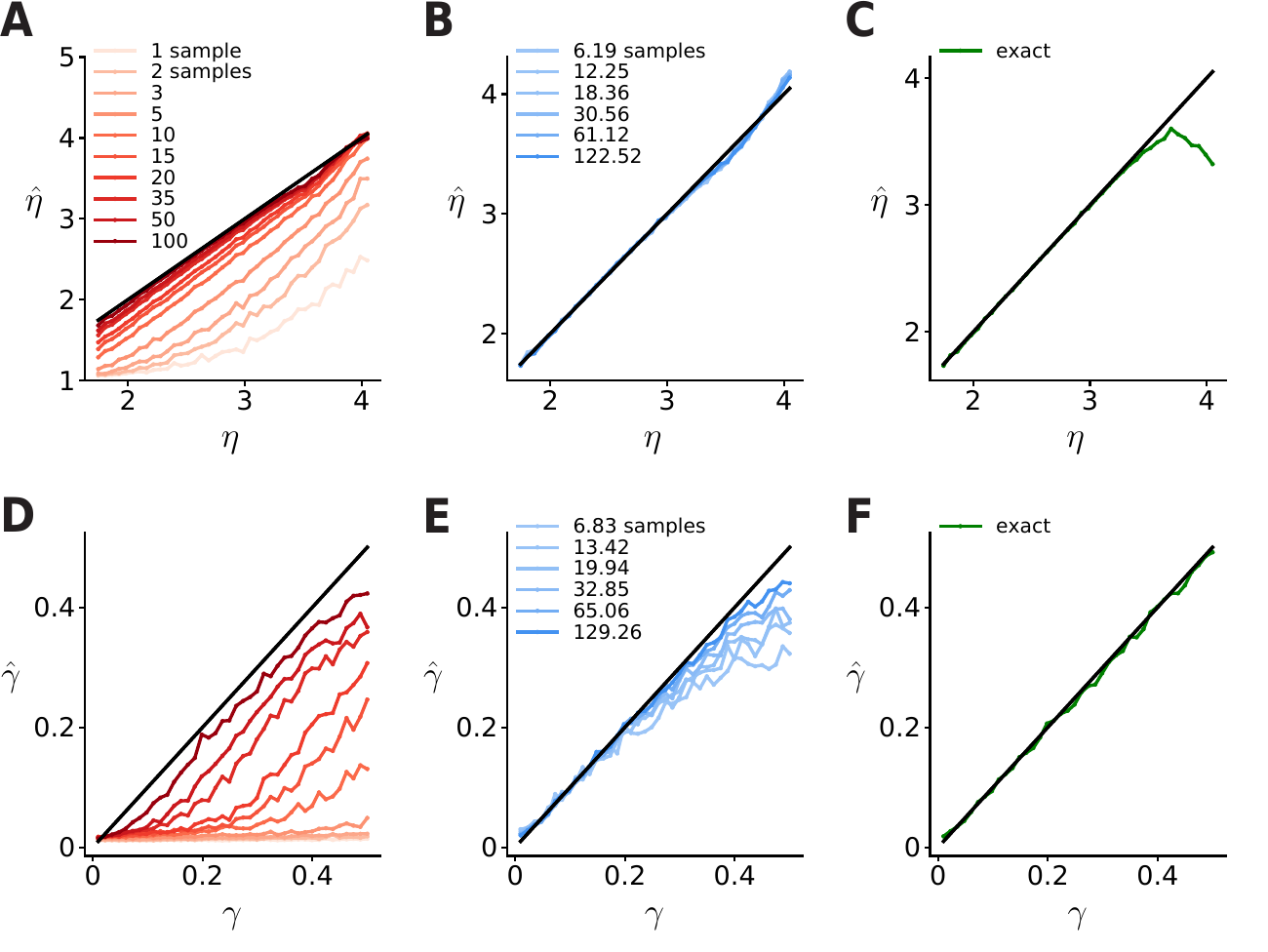}
  \vspace{-0.5em}
  \caption{Same as Figure~\ref{fig:psycho_complete_results}, for the change localization model. Fixed sampling is substantially biased for both the measurement noise $\eta$ and the lapse rate $\lapse$, whereas IBS is accurate for $\eta$ and biased for $\lapse$, but still much less biased than fixed sampling.}
\label{fig:vstm_complete_results}
\end{figure}

We report in Figure~\ref{fig:vstm_complete_results} the parameter recovery results for fixed sampling, inverse binomial sampling and `exact' methods for the change localization task, for all tested number of samples $M$ and IBS repeats $R$. For this task, the exact method relies on numerical integration.

\subsection{Four-in-a-row game}
\label{sec:fourinarow_details}

The four-in-a-row game model parameters are the value noise $\eta \equiv \log \sigma$, the pruning threshold $\thresh$, and the feature dropping rate $\delta$, with bounds defined in Table \ref{tab:fourinarow_params}.
We use the same procedure and settings for BADS as for the orientation discrimination task (see Section \ref{sec:ori_details}), unless noted otherwise. For IBS, we use an early-stopping threshold of $\L_\text{lower} = -N \log 20$, and due to computational cost we use only $R \in \{1,2,3\}$. For fixed sampling we consider $M \in \{1,2,3,5,10,15,20,35,50,100\}$. We have no expression for the likelihood of the four-in-a-row game model, not even in numerical form, so there is no `exact' method.

\begin{table}[ht]
\caption{Parameters and bounds of the four-in-a-row game model.}
\label{tab:fourinarow_params}
 \vspace{.5em}
\begin{center}
\begin{tabular}{|cl|cc|cc|}\hline
Parameter & Description & LB    & UB  & PLB  & PUB \\
\hline 
$\eta \equiv \log \sigma$ & Value noise  & $\log 0.01$ & $\log 5$ & $\log 0.2$ & $\log 3$ \\
$\thresh$ & Pruning threshold  & $0.01$ & $10$ & $1$ & $10$ \\
 $\delta$ & Feature dropping rate  & $0$ & $1$  & $0$ & $0.5$ \\\hline
\end{tabular}
\end{center}
\end{table}

We run BADS $8$ times, with starting values of $\eta \in\{-0.707,0.196\}$, $\thresh \in \{4,7\}$ and $\delta \in \{0.167, 0.333 \}$. For data generation, we set as baseline parameter vector $\eta = \log 1$, $\thresh = 5$ and $\delta = 0.2$ and for each parameter we select $40$ parameter vectors linearly spaced in the plausible range for that parameter (as per Table \ref{tab:fourinarow_params}), while keeping the other two parameters at their baseline value. Again, we generate $100$ data sets for each such parameter combination.

We fixed the other parameters of the model to typical values found in the previous study \citep{van2016people}, namely $w_\text{center} = 0.60913$, $w_\text{connected 2-in-a-row} = 0.90444$, $w_\text{unconnected 2-in-a-row} = 0.45076$, $w_\text{3-in-a-row} = 3.4272$, $w_\text{3-in-a-row} = 6.1728$, $C_\text{act} = 0.92498$, $\gamma_\text{tree}=0.02$, $\lambda=0.05$. The $w_i$ are the weights of features $f_i$ in the value function, briefly $f_\text{center}$ values pieces near the center of the board, the other features count the number of times certain patterns occur on the board (see \citealp{van2016people} for the specific patterns). $C_\text{act}$ is a parameter which scales the value of features belonging to the active or passive player. The parameter $\gamma_\text{tree}$ is inversely proportional to the size of the tree built by the algorithm, and $\lambda$ is the lapse rate, that is the probability of a uniformly random move among the available squares (note that for the other models we denoted lapse rate as $\lapse$; here we use the variable naming from \citealp{van2016people}). See \citet{van2016people} for more details about the model and its parameters. 

\subsubsection*{Complete parameter recovery results}

\begin{figure}[htp]
	\centering
	\includegraphics[width=5.2in]{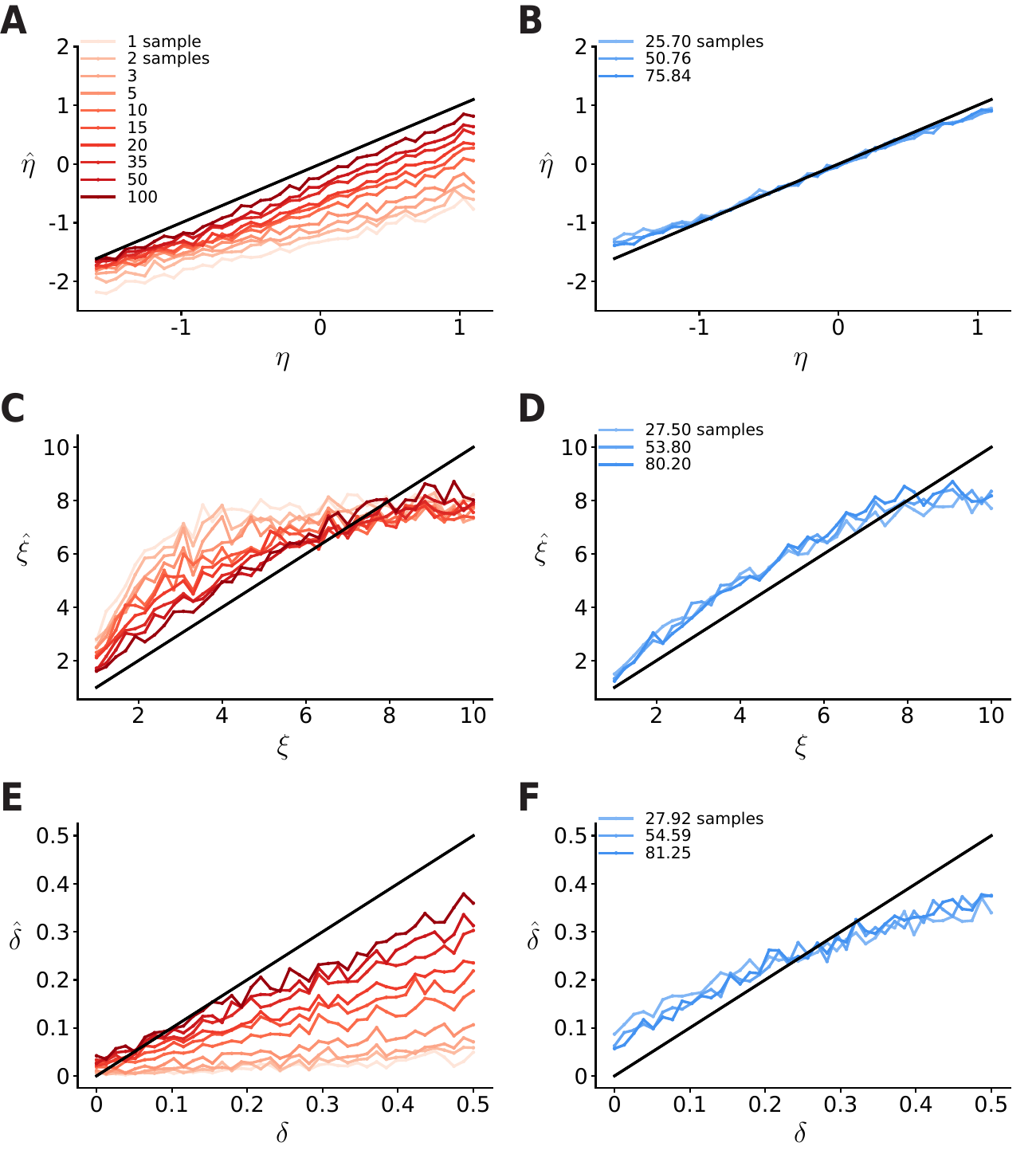}
	\vspace{-0.5em}
\caption{Same as Figure~\ref{fig:psycho_complete_results}, for the four-in-a-row task. For this model, we do not have an exact log-likelihood formula or numerical approximation, so we only show fixed sampling and IBS. Overall, fixed sampling has substantial biases in its estimation of $\eta$ and $\delta$ and a smaller bias in estimating $\thresh$. IBS has almost no bias for $\eta$ and only a small bias for $\thresh$ and $\delta$.}
\label{fig:fourinarow_complete_results}
\end{figure}

We report in Figure~\ref{fig:fourinarow_complete_results} the parameter recovery results for fixed sampling and inverse binomial sampling for the 4-in-a-row task, for all tested number of samples $M$ and IBS repeats $R$. For this task, there is no `exact' method to evaluate the log-likelihood.

\section{Improvements of IBS and further applications}
\label{sec:improvements}

\subsection{Early stopping threshold}
\label{sec:threshold}

One downside of inverse binomial sampling is that the computational time it uses to estimate the log-likelihood is of the order of $\frac{1}{p}$, which is equal to $\exp\left(-\log p\right) = \exp\left(-\L \right)$. In other words, IBS spends exponentially more time on estimating log-likelihoods of poorly-fitting models or bad parameters. This implies that an optimization algorithm that uses IBS allocates more computational resources to estimating the objective function $\L\left(\vtheta\right)$ for parameter vectors $\vtheta$ where the objective is low. However, the value of the objective at such poor parameter vectors are unlikely to affect its estimate of the location or value of the maximum, so the optimizer (BADS in our case) is wasting time. It may be possible to develop optimization algorithms that take into account the exponentially large cost of probing points where the objective function is low, but we can circumvent the problem by amending IBS with a criterion that stops sampling when it realizes that $\hat{\L}\left(\vtheta\right)$ will be low. 

\begin{figure}[htp]
	\centering
	\includegraphics[width=5.2in]{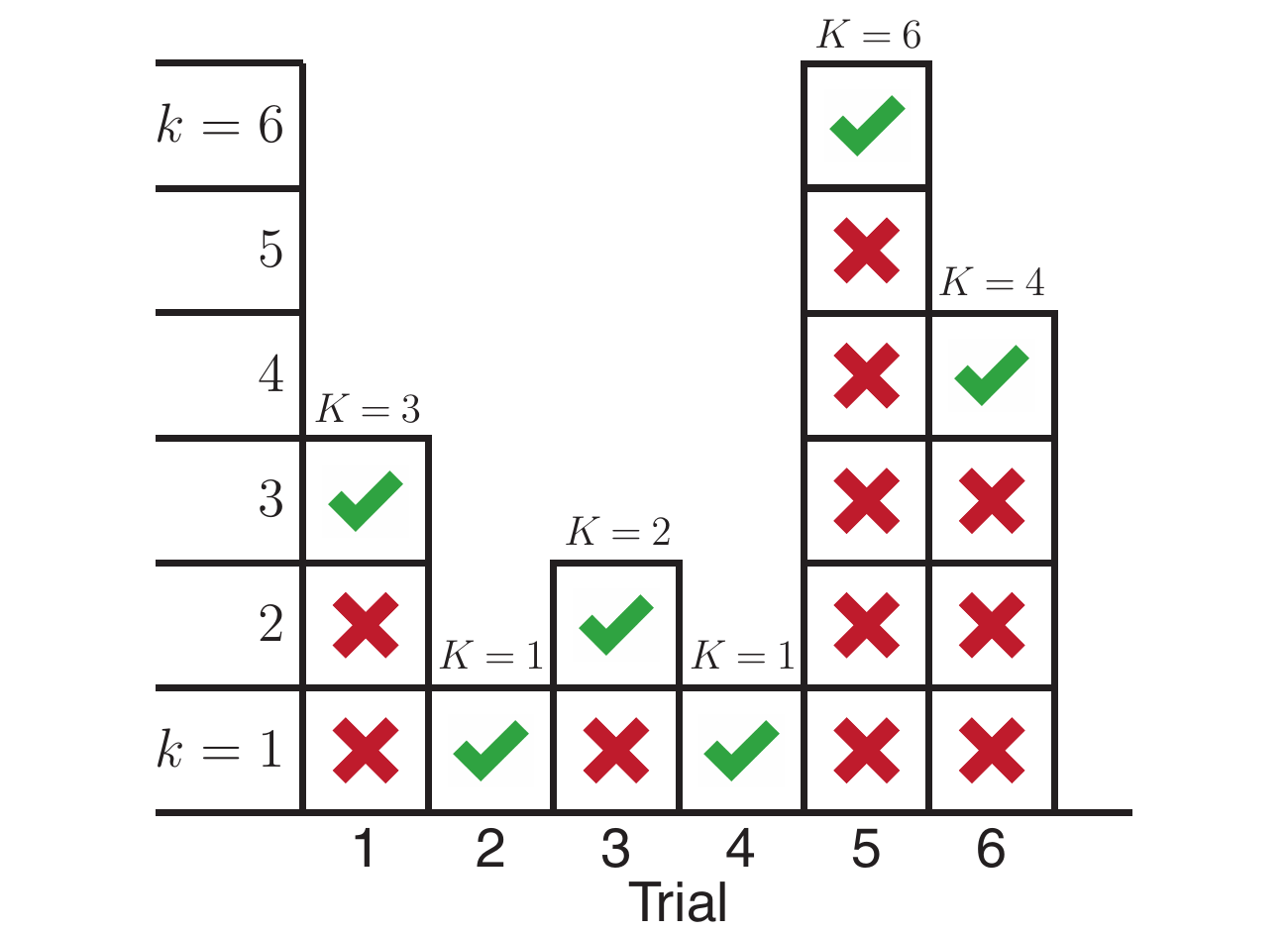}
	\vspace{-0.5em}
\caption{Graphical illustration of the two methods to implement IBS with multiple trials, in this case $N = 6$. In this figure, each column represents a trial, each box above the trial number represents a successive sample from the model from that trial, with red crosses for samples that do not match the participant's response (`misses') and green checkmarks for ones that do (`hits'). Above each column, we indicate $K$, the number of samples until a hit. For trials $2$ and $4$, $K=1$ so $\Libs=0$. The most obvious implementation of multi-trial IBS is `columns-first', to sample model responses for each trial until a hit, and only then move to the next trial. However, a more convenient sampling method is `rows-first', and sample one response for each trial with $k=1$, then one response for each trial with $k=2$, excluding trials $2$ and $4$ since the first sample was a hit, and continue increasing $k$ until all trials reach a hit. This method allows for early stopping and a parallel processing.}
\label{fig:sampling_scheme}
\end{figure}

In Section~\ref{sec:implementation}, we introduced a basic implementation of IBS for estimating the log-likelihood of multiple trials, by sequentially computing the log-likelihood of each trial. However, another way to implement multi-trial IBS (a `parallel' implementation) is to draw one sample from the simulator model for each trial, then set $K_i=1$ for each trial where the sample matches the participant's response. For all other trials, draw a second sample from the model, and if that matches the response, set $K_i=2$. Finally, repeat this process until no more trials remain. We illustrate this sampling scheme graphically in Figure~\ref{fig:sampling_scheme}. 

After each iteration, we then compute
\begin{equation}
\hat{\L}_K=-\sum_{i\in\mathcal{I}_{\text{match}}}\sum_{k=1}^{K_i-1}\frac{1}{k} - N_{\text{remaining}} \sum_{k=1}^{K-1}\frac{1}{k}
\end{equation}
where $K$ is the iteration number, $\mathcal{I}_{\text{match}}$ is the set of trials where we found a matching sample and $N_{\text{remaining}}$ is the number of remaining trials. This value $\hat{\L}_K$ is decreasing and by construction converges to $\displaystyle\sum_{i=1}^N\hat{\L}_{i,\text{IBS}}$ as $K\rightarrow\infty$. Therefore, whenever $\hat{\L}_K$ falls below a lower bound $\L_{\text{lower}}$, we are guaranteed that $\Libs$ will be below that bound too. When it does, we stop sampling and return $\L_{\text{lower}}$ as estimate of $\L\left(\vtheta\right)$. This does introduce bias into the estimate, but since we bound the \emph{total} log-likelihood, the bias will be exponentially small in $N$ as long as the true value $\L\left(\vtheta\right)$ is adequately larger than $\L_{\text{lower}}$. 

In practice, we recommend using as lower bound the log-probability of the data under a `chance' model, which assigns uniform probability to each possible response on each trial, and should be a poor model of the data. In the orientation discrimination and change localization examples from Sections \ref{sec:ori} and \ref{sec:change}, the log-likelihood of a chance model is $-N \log 2$ and $- N \log 6$, respectively. For the 4-in-a-row game presented in Section \ref{sec:fourinarow} the log-likelihood of chance depends on the number of pieces on each board position; we chose an average value such that $\L_\text{lower} = - N\log 20$. This new sampling scheme has an additional advantage: since on each iteration we independently sample from the generative model on multiple trials, we can potentially run these computations in parallel. 

\subsection{Reducing variance by trial-dependent repeated sampling}\label{sec:repeated_sampling}

As we saw in Section \ref{sec:multifidelity}, a simple method to improve the estimate of IBS is to run the estimator multiple times and average the results. Repeated sampling will preserve the zero bias but reduce variance inversely proportional to the number of repeats $R$.

We can further improve the estimator by varying the number of repeats $R_i$ between trials, for $1 \le i \le N$, and define
\begin{equation} \label{eq:multireps}
\Libsrep{\R}=\sum_{i=1}^N\frac{1}{R_i}\sum_{r=1}^{R_i}\hat{\L}_i^{(r)},
\end{equation}
where $\R$ is a vector of positive integers with elements $R_i$, and $\hat{\L}_i^{(r)}$ denotes the outcome of the $r$-th run of IBS on trial $i$. This estimator is unbiased regardless of the choice of $\R$ (as long as $R_i>0$ for all trials), and we can analytically compute both its variance and expected number of samples (see \eq~\ref{eq:lagrange}).

We can then ask what is the best allocation of repeats $R_i$ that minimizes the variance of the estimator in \eq~\ref{eq:multireps} such that the expected total number of samples does not exceed a fixed budget $\budget$. This defines the following constrained optimization problem,
\begin{equation}\label{eq:lagrange}
\R^* = \arg \min_{R_i,R_2\ldots,R_N}\left\{\left.\frac{1}{N}\sum\limits_{i=1}^N\frac{\text{Li}_2(1-p_i)}{R_i}\right\lvert\sum\limits_{i=1}^N\frac{R_i}{p_i}\le \budget\right\}
\end{equation}
where we used \eq~\ref{eq:ibsvariance} for the variance of the IBS estimator.

\begin{figure}[htp]
	\centering
	\includegraphics[width=5.2in]{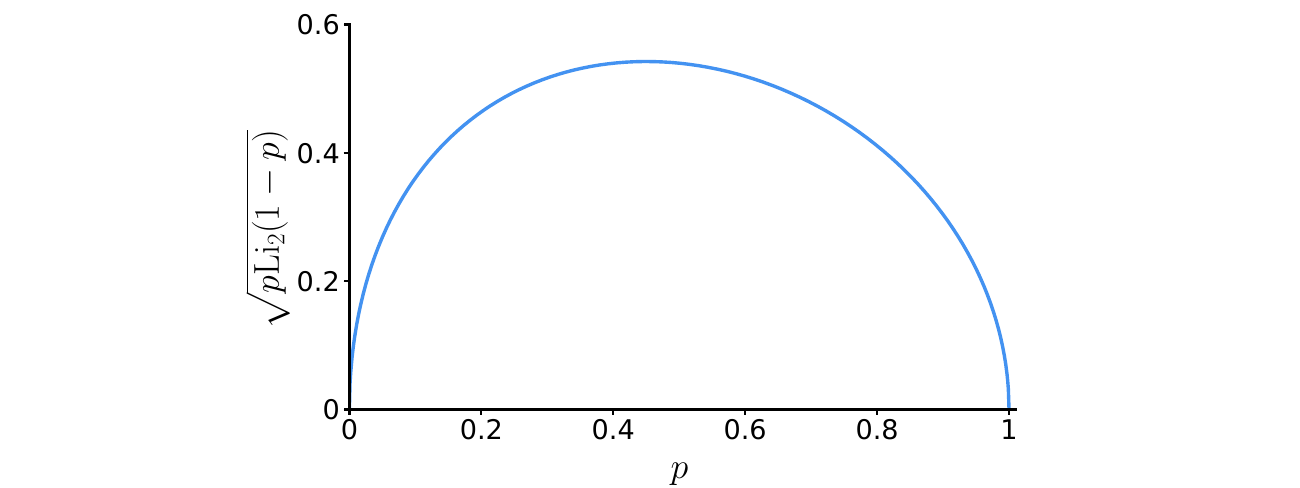}
	\vspace{-0.5em}
\caption{Graph of $\sqrt{p\text{Li}_2(1-p)}$, which is proportional to the optimal number of repeats for a trial with likelihood $p$ (see equation~\ref{eq:Ropt}). We observe that the optimal allocation of computational resources entails repeated sampling for trials with $p\approx\frac{1}{2}$ and to avoid $p\approx 0$ or $p\approx 1$.}
\label{fig:repeats}
\end{figure}

Assuming that the $R_i$ take continuous values, we can solve the optimization problem in \eq~\ref{eq:lagrange} exactly using a Lagrange multiplier, and find the following closed-form expression for the optimal number of repeats per trial,
\begin{equation}\label{eq:Ropt}
R_i^*=\budget \left(\sum\limits_{j=1}^N\sqrt{\frac{\text{Li}_2(1-p_j)}{p_j}}\right)^{-1}\sqrt{p_i\text{Li}_2(1-p_i)}.
\end{equation}
According to \eq~\ref{eq:Ropt}, the optimal choice of repeats entails dividing the budget $\budget$ across trials, where trial $i$ is allocated repeats proportional to $\sqrt{p_i\text{Li}_2(1-p_i)}$. We plot this function in Figure~\ref{fig:repeats} and see that, to minimize variance, we should allocate resources primarily to trials where $p_i$ is close to $\frac{1}{2}$ and avoid trials where $p_i\approx 1$ (since the variance of IBS is already small for those trials) or where $p_i\approx 0$ (since those utilize a larger share of the budget). 

We can also calculate exactly the fractional increase in precision (inverse variance) when using the optimal choice of repeats vector $\R^*$, compared to a constant $R$ which divides the budget equally across trials,
\begin{equation}\label{eq:Ropt_gain}
\frac{\text{Var}\left[\Libsrep{R}\right]}{\text{Var}\left[\Libsrep{\R^*}\right]}=\left(\sum\limits_{i=1}^N\sqrt{\frac{\text{Li}_2(1-p_i)}{p_i}}\right)^2\times\left(\sum\limits_{i=1}^N\text{Li}_2(1-p_i)\right)^{-1}\times\left(\sum\limits_{i=1}^N\frac{1}{p_i}\right)^{-1}.
\end{equation}
This equation implies that the gain in precision from this method depends on the distribution of $p_i$ across trials. If $p_i \sim \text{Uniform[0,1]}$ and $N=500$, the median precision gain is $1.584$ and the inter-quartile range (IQR) is $1.375$ to $2.090$. Note that the gain is always greater than $1$, unless $p_i$ is constant across trials. 

\subsubsection*{Practical implementation of trial-dependent repeated sampling}

In practice, \eq~\ref{eq:Ropt} cannot be applied directly, as we have treated $R_i$ as continuous variables, but the number of times to repeat IBS on a given trial has to be an integer. Additionally, the method is only unbiased if $R_i$ is at least $1$ for each trial $i$. Therefore, we can convert $R_i^*$ to integers by rounding up to the nearest integer. This method will make our solution approximate, and reduce the gain in precision, but it is still better than uniform repeats for uniformly distributed $p_i$ (median: $1.567$, IQR: $1.374$-$2.002$).

The derivation above has another, more fundamental problem. Computing $R_i^*$ requires knowledge of $p_i$ on each trial, which we do not have. While we could try and learn the allocation of $\R^*$ as a function of $\vtheta$ in some adaptive way, in practice we recommend the following simple scheme:
\begin{enumerate}
\item Choose a default parameter vector $\vtheta_0$, and run IBS with a large number of repeats (e.g, $R = 100$) to estimate the (log)-likelihood of the model on each trial.
\item Compute the optimal repeats $R_i^*$ given the estimated likelihoods $\hat{p}_i$ and a total budget of expected samples $\budget$ per likelihood evaluation, and round up. 
\item Run IBS with those fixed repeats per trial on each iteration of the optimization algorithm.
\end{enumerate}
This approach implicitly assumes that the log-likelihood will be correlated across trials between the generative model with parameter vector $\vtheta_0$ and any other vector $\vtheta$ probed by the optimization algorithm. This is usually the case, since low-probability trials are often those where something unexpected occurred (e.g., the participant of a behavioral experiment lapsed or otherwise made an error). In our experience, this scheme considerably reduces the variance of IBS for a given computational time budget.

\subsection{Bayesian inference with IBS}
\label{sec:bayesian}

While the main text focused on maximum-likelihood estimation, the unbiased log-likelihood estimates provided by IBS can also be used to perform Bayesian inference of posterior distributions over model parameters. We describe here a few possible approaches to approximate Bayesian inference with IBS.

\subsubsection*{Markov Chain Monte Carlo}

Markov Chain Monte Carlo (MCMC; see e.g. \citealp{brooks2011handbook}) is a powerful class of algorithms that allows one to sequentially sample from a target probability density which is known up to a normalization constant (e.g., the joint distribution). A popular form of MCMC is known as Metropolis-Hastings (MH; \citealp{hastings1970monte}), which explores the target distribution by drawing a sample from a `proposal distribution' centered on the last sample (e.g., a multi-variate Gaussian).
MH `accepts' or `rejects' the new sample with an acceptance probability that depends on the value of the target density at the proposed and at the last point. In case of acceptance, the new point is added the sequence of samples; otherwise, the last sample is repeated in the sequence. Under some conditions, the MH algorithm produces a (correlated) sequence of samples that are equivalent to draws from the target density. Crucially, and somewhat surprisingly, the MH algorithm is still valid (that is, produces a valid sequence) if one performs the comparison with a noisy but unbiased estimate of the target density as opposed to using the exact density \citep{andrieu2009pseudo}.

One problem here is that IBS provides an unbiased estimate of the log-likelihood (and thus of the log target density); not of the likelihood. However, since the IBS estimates of the log-likelihood are nearly-exactly normally distributed (see Section \ref{sec:variance}), the distribution of the likelihood is log-normal. Thus, we can apply what is known as a `convexity correction' and compute a (nearly) unbiased estimate of the likelihood $\hat{\ell}(\vtheta)$ by calculating the expected value of a log-normal variable, that is
\begin{equation} \label{eq:convexity}
\hat{\ell}(\vtheta) = \exp\left(\hat{\L}(\vtheta) + \frac{1}{2}\text{Var}\left[\hat{\L}(\vtheta)\right]\right).
\end{equation}
\eq~\ref{eq:convexity} can be easily evaluated with IBS, using the expression for the variance of the IBS estimator (\eq~\ref{eq:ibsvarest}).

\subsubsection*{Variational inference}

An alternative class of approximate inference methods is based on \emph{variational inference} (VI; \citealp{jordan1999introduction}). The goal of VI is to approximate the intractable posterior distribution with a simpler distribution $q(\vtheta)$ belonging to a chosen parametric family. A common choice is a multivariate normal with diagonal covariance (known as \emph{mean field approximation}); but other choices are possible too. VI selects the `best' approximation $q(\cdot)$ that minimizes the Kullback-Leibler divergence with the true posterior, or equivalently maximizes the following variational objective,
\begin{equation} \label{eq:elbo}
\mathcal{E}\left[q\right] = \mathbb{E}_{\vtheta \sim q(\cdot)}\left[\L(\vtheta)\right] + \mathcal{H}\left[q\right],
\end{equation}
where $\mathcal{H}\left[q\right]$ is the entropy of $q(\cdot)$, which we assume can be computed analytically or numerically. Crucially, we can obtain an unbiased estimate of the first term in \eq~\ref{eq:elbo} (the expected log joint) with IBS, as we have seen in Section \ref{sec:applications}. The optimization of the variational objective can then be performed directly with derivative-free optimization methods (such as BADS), or via a technique that produces unbiased estimates of the gradient combined with variance-reduction tricks, called \emph{black-box variational inference} \citep{ranganath2014black}.

\subsubsection*{Gaussian process surrogate methods}

One issue with the approximate inference methods described above is that they require a large (possibly, \emph{very} large) number of likelihood evaluations to converge. Thus, these approaches are unfeasible if the generative model is somewhat computationally expensive, as it is often the case. An alternative family of methods designed to deal with expensive likelihoods builds a Gaussian process approximation (a surrogate) of the log joint distribution, and uses it to actively acquire further points in a smart way, similarly to the approach of Bayesian optimization \citep{kandasamy2015bayesian,acerbi2018variational,jarvenpaa2019parallel}. However, unlike Bayesian optimization, the goal here is not to \emph{optimize} the target function, but instead to build an accurate approximation of the posterior distribution, with as few likelihood evaluations as possible. 

IBS is particularly suited to be used in combination with Gaussian process surrogate methods as it provides both an unbiased estimate of the log-likelihood, and a calibrated estimate of the uncertainty in each measurement, which can be used to inform the Gaussian process observation model. The development of Gaussian process surrogate methods is an active and very promising area of research. A recent example is Variational Bayesian Monte Carlo (VBMC; \citealp{acerbi2018variational,acerbi2019exploration}), a technique that naturally combines Gaussian process surrogate modeling with variational inference thanks to Bayesian quadrature \citep{ohagan1991bayes}. Conveniently, VBMC returns both an approximate posterior distribution and an estimate of the model evidence, which can be used for model comparison. \citet{acerbi2020variational} showed that VBMC, combined with IBS and modified to deal with noisy log-likelihood evaluations, performs very well on a variety of models from computational and cognitive neuroscience.

\end{document}